\documentclass{article} 
\usepackage{iclr2027_conference,times}

\usepackage{amsmath,amsfonts,bm}

\def\eqref#1{equation~\ref{#1}}

\def\1{\bm{1}}

\DeclareMathAlphabet{\mathsfit}{\encodingdefault}{\sfdefault}{m}{sl}
\SetMathAlphabet{\mathsfit}{bold}{\encodingdefault}{\sfdefault}{bx}{n}

\usepackage{hyperref}
\usepackage{url}
\usepackage{graphicx}
\usepackage{booktabs}
\usepackage{amssymb}
\usepackage{amsthm}
\usepackage{xcolor}
\usepackage[table]{xcolor}
\usepackage{soul}
\usepackage{multirow}
\usepackage{wrapfig}

\newtheorem{lemma}{Lemma}

\title{CentriQ: Calibration-Free Quantization of Diffusion Transformers via Exact Mean Centering }

\iclrfinalcopy
\author{Nataša Jovanović $^{ 1, 2}$\hspace{0.5cm}Mathieu Salzmann $^{ 2}$\hspace{0.5cm}Saqib Javed $^{ 2}$   \\
$^{ 1}$ Tenstorrent\hspace{0.5cm}$^{ 2}$ EPFL \\
}

\begin{document}

\newcommand{\ph}{\textcolor{red}{\texttt{??}}}      
\newcommand{\cmark}{\checkmark}
\newcommand{\xmark}{\ensuremath{\times}}
\newcommand{\pub}{\textsuperscript{\dag}}            

\maketitle

\begin{abstract}

Diffusion transformers (DiTs) achieve state-of-the-art image generation, but their sampling cost limits deployment. Quantizing both weights and activations to 4 bits reduces this cost, yet existing methods fall short in one of two ways. Calibration-based methods are tied to a specific checkpoint and prompt distribution, whereas data-free Hadamard rotation, effective for LLMs, loses quality on DiTs. We show that this loss has a structural cause. Adaptive layer-norm conditioning adds a per-token mean to the activations, and at the widths of the evaluated DiTs, the Hadamard rotations used by data-free methods cannot spread this mean uniformly across coordinates. A single dominant direction therefore survives the rotation and sets the quantization range. We introduce CentriQ, a calibration-free quantizer that centers each token before rotation and restores the mean exactly through a rank-1 full-precision branch, so that per-token scales follow in closed form without data. Weights are fitted under a robust $\ell_p$ objective that tracks the dense mode of each group and discounts heavy tails. Across three DiTs, CentriQ matches the quality of calibrated SVDQuant at 4 bits, whereas calibration-free weight quantizers with plain per-token activation quantization collapse or degrade substantially. CentriQ outperforms the strongest calibration-free method reported to date at 2-bit weights. It is also the first calibration-free method to retain usable image quality at 2-bit activations.

\end{abstract}

\section{Introduction}

Diffusion transformers (DiTs)~\citep{peebles2023scalablediffusionmodelstransformers} underlie state-of-the-art text-to-image models, but sampling them is expensive. Each image requires many denoising steps, and each step is dominated by the linear layers of the transformer blocks. Unlike autoregressive LLM decoding, which at small batch sizes is limited by memory bandwidth, a DiT forward pass is compute-bound. Weight-only quantization (e.g. W4A16) therefore reduces memory but not latency, because the low-bit weights are upcast to 16 bits before the matrix multiply, so activations must be quantized as well.

The post-training methods that do reach W4A4 on diffusion models all rely on calibration. A prompt set is run through the model to fit smoothing factors, clipping ranges, rotations, or error-compensating weight updates to the activation statistics of that specific model. SVDQuant~\citep{li2025svdquant}, for instance, migrates activation outliers into the weights using smoothing factors computed from COCO caption prompts, absorbs the magnified weight outliers in a 16-bit low-rank branch, and in its final version also applies GPTQ to the residual.

Calibration has costs beyond its compute budget. It makes the quantized model depend on a dataset, it must be repeated for every model, fine-tune, and quantization configuration, and its quality depends on how well the calibration prompts and timesteps match the deployment distribution. Calibration-free methods that reach four-bit activations on image DiTs have appeared only recently (Sec.~\ref{sec:related}), and they evaluate on generation benchmarks that do not measure fidelity to the full-precision model.

Two properties of DiT activations make the calibration-free setting difficult. First, adaptive layer normalization shifts and rescales every token with conditioning-dependent parameters, so each activation vector carries a per-token mean. A normalized Hadamard rotation, the standard tool for spreading outliers, maps this component onto one or a few coordinates, and the resulting outlier sets the quantization range for the whole vector. Second, only regular Hadamard matrices, whose rows have equal sums, spread this component uniformly, and these exist only at orders $n = 4u^2$. None of the hidden widths of PixArt-$\Sigma$ (1152), FLUX.1 (3072), and SANA-1.6B (2240) has this form, so rotation methods operate through block-structured or Kronecker constructions, which inherit the mean problem.

We present CentriQ, a calibration-free W4A4 post-training quantization method for DiTs that addresses both properties directly. It subtracts the per-token mean before rotation, quantizes only the residual, and restores the mean exactly as a rank-1 term in the layer output, so the quantization range no longer depends on the offset. The centered activation is rotated by a deterministic factorized Hadamard matrix, and weights are quantized to grouped INT4 under a robust $\ell_p$ ($p<1$) objective, so neither path uses calibration data.

Our contributions can be summarized as follows:

\begin{itemize}
\item We show that adaLN places a per-token mean in DiT activations, which a Hadamard rotation concentrates onto one or a few coordinates that set the quantization range (Figs.~\ref{fig:dc-energy} and~\ref{fig:sorted-mag}).
\item We show that after centering, the rotated operand, its per-token scale, and its quantization error contain no term that depends on the row sums of the rotation (Sec.~\ref{sec:activation}, App.~\ref{sec:analysis}). The choice of rotation is thereby decoupled from the mean, so CentriQ mixes globally with factorized Hadamard matrices and restores the mean exactly as a rank-1 term in the GEMM epilogue.
\item CentriQ uses no calibration data and reaches FID and ImageReward that match calibration-based methods at W4A4 on three DiT families. Furthermore, it surpasses the strongest reported calibration-free method at W2A4 and is the first calibration-free method to hold image quality at W4A2 (Tab.~\ref{tab:main-w4a4}, Tab.~\ref{tab:lowbit-flux}, Tab.~\ref{tab:geneval-flux}).
\end{itemize}

Our code and models will be made publicly available.
\vspace{-0.3cm}
\section{Related work}
\label{sec:related}

\paragraph{PTQ for diffusion models.}
Work on U-Net diffusion models showed that activation statistics drift with the timestep~\citep{li2023qdiffusion,shang2023ptq4dm,he2023ptqd,so2023tdq,huang2024tfmq}. Methods for diffusion transformers inherit this diagnosis and absorb the drift with calibration: PTQ4DiT~\citep{wu2024ptq4dit} weights salient channels by timestep, Q-DiT~\citep{chen2025qdit} and ViDiT-Q~\citep{zhao2025viditq} use dynamic per-timestep or per-token activation scales, DiTAS~\citep{dong2025ditas} aggregates smoothing factors across timesteps, LRQ-DiT~\citep{yang2025lrqdit} chooses rotations per layer from a calibrated metric, and AdaTSQ~\citep{zhang2026adatsq} allocates bit-widths along the trajectory with Fisher-weighted Hessian calibration. The fitted statistics belong to one checkpoint and sampler, so these methods must be refit when the model or guidance changes. Sampler-side corrections~\citep{he2023ptqd,ryu2026qdrift} act after quantization and are complementary to any quantizer, whereas training-based approaches~\citep{he2024efficientdm,sui2024bitsfusion,yang2025robuq,huang2026qvgen} fall outside the post-training setting.

\paragraph{Rotation and incoherence.}
LLM quantization spreads activation outliers across channels by rotating the residual stream, with random orthogonal matrices in QuIP~\citep{chee2023quip}, randomized Hadamard transforms in QuIP\#~\citep{tseng2024quipsharp}, and Hadamard rotations folded into the weights in QuaRot~\citep{ashkboos2024quarot}. Learned rotations~\citep{liu2025spinquant,lin2024duquant,sun2025flatquant,hu2025ostquant} recover accuracy that a fixed Hadamard loses, but they are fit on calibration data. All of these assume a Hadamard matrix of the model width, which the Sylvester construction gives only for powers of two, so other widths use a Kronecker product $H_{2^k}\otimes H_m$ with a small hand-picked $H_m$.

\paragraph{Mean removal.}
SliceGPT~\citep{ashkboos2024slicegpt} and QuaRot fold LayerNorm mean subtraction into the preceding weights, which is exact in LLMs. In a DiT, the adaLN modulation depends on the timestep and prompt and makes the mean vary per token, so this static fold does not apply. HadaNorm~\citep{federici2025hadanorm} subtracts a per-channel mean computed over tokens at inference and applies SmoothQuant-style per-channel scales, fitted on a calibration set, before a Hadamard transform, and reports W4A4 on PixArt-$\Sigma$. CentriQ removes the per-token mean instead and requires no calibration data. To our knowledge, no public implementation of HadaNorm is available, which precludes a direct comparison under our protocol. ConvRot~\citep{huang2025convrot} spreads the mean with group-wise regular Hadamard transforms of order $4^k$, which mix only within each group, whereas CentriQ removes the mean and mixes globally with factorized Hadamard matrices.

\paragraph{Calibrated W4A4 for DiTs.}
SVDQuant~\citep{li2025svdquant} was the first to reach reasonable 4-bit weights and activations on image DiTs, by keeping a low-rank branch in 16 bits after SmoothQuant-style migration~\citep{xiao2023smoothquant} fit on captioned images, with fused kernels in Nunchaku. DiRotQ~\citep{sharify2026dirotq} preserves a calibration-derived PCA subspace of the activations at higher precision and quantizes weights with GPTQ~\citep{frantar2023gptq}, improving over SVDQuant on PixArt-$\Sigma$ under the MJHQ-30K protocol, which is not directly comparable with the MJHQ-5k protocol used here. HQ-DiT~\citep{liu2024hqdit} reaches 4-bit operands with a floating-point grid.

\paragraph{Calibration-free quantizers.}
On the weight side, RTN, NF4~\citep{dettmers2023qlora}, HQQ~\citep{badri2023hqq}, and SINQ~\citep{muller2025sinq} operate on the weight tensor alone. NF4 assumes near-Gaussian weights, HQQ fits the dense mode under an $\ell_p$ objective with $p<1$, and SINQ balances a second scale axis with Sinkhorn iterations. Calibration-free 4-bit activations on DiTs are recent. DVD-Quant~\citep{li2026dvdquant} achieves W4A4 for video DiTs with grid refinement, rotated quantization, and adaptive bit switching. ConvRot achieves it on FLUX-family models with group-wise rotation, NVFP4 weights, and per-tensor FP8 activation scales. OrbitQuant~\citep{lee2026orbitquant} quantizes weights and activations in a randomly permuted block-Hadamard basis with a Lloyd--Max codebook derived from the post-rotation marginal~\citep{zandieh2025turboquant}, and reports GenEval on FLUX.1 and Z-Image-Turbo down to W2A4. Neither image method removes the token mean, both mix within blocks whose order divides the width, and both evaluate on generation benchmarks that do not measure fidelity to the full-precision model.

\paragraph{Position of this work.}
CentriQ is calibration-free on both axes and produces integer W4A4 operands for every transformer-block projection. We evaluate it on PixArt-$\Sigma$, FLUX.1-schnell, and SANA-1.6B with FID, ImageReward, LPIPS, and PSNR against the 16-bit reference on MJHQ, and compare against the published results of calibrated SVDQuant. To our knowledge it is the first calibration-free integer W4A4 method evaluated for fidelity across three image-DiT architecture families.

\section{Preliminaries}
\label{sec:prelim}

A linear layer computes $\mathbf{y}_t = W\mathbf{x}_t + \mathbf{b}$, where $W \in \mathbb{R}^{d_{\mathrm{out}}\times n}$ and $\mathbf{x}_t \in \mathbb{R}^{n}$ is the activation of token $t$. We write $\mathbf{1}$ for the all-ones vector, $\|\cdot\|_\infty$ for the largest coordinate magnitude, and a hat for a quantized tensor.

\paragraph{Quantizers.}
Activations are quantized online to a symmetric $b$-bit grid with one 16-bit scale per token.
\begin{equation}
Q_a(\mathbf{x}_t) = \alpha_t \cdot \mathrm{clip}\!\big(\mathrm{round}(\mathbf{x}_t/\alpha_t),\,-(2^{b-1}-1),\,2^{b-1}-1\big), \qquad \alpha_t = \frac{r_t}{2^{b-1}-1},
\label{eq:act-quant}
\end{equation}
where the range $r_t$ is either $\|\mathbf{x}_t\|_\infty$ or a high percentile of $|\mathbf{x}_t|$, computed from the current token at inference. Because one scale serves all $n$ coordinates, a single large coordinate coarsens the grid for the whole token. Weights are quantized to an asymmetric $b$-bit grid over groups of $g$ consecutive input columns,
\begin{equation}
\widehat{W}_{g} = \boldsymbol{\sigma}\,(Q_{g} - \mathbf{z}), \qquad Q_{g} \in \{0,\dots,2^{b}-1\}^{d_{\mathrm{out}}\times g},
\label{eq:weight-quant}
\end{equation}
with a per-(row, group) scale $\boldsymbol{\sigma}$ and shift $\mathbf{z}$ stored in 16 bits. Round-to-nearest (RTN) sets both from the group minimum and maximum, an $\ell_\infty$ criterion, and an $\ell_2$-optimal scale is likewise dominated by the largest entries of the group, which motivates the criterion of Sec.~\ref{sec:weights}.

\paragraph{Hadamard rotation.}
A Hadamard matrix $H_n \in \{\pm 1\}^{n\times n}$ satisfies $H_nH_n^{\top} = nI$, so $H = \tfrac{1}{\sqrt{n}}H_n$ is orthogonal and can be folded into the weight without changing the layer, $W\mathbf{x} = (WH^{\top})(H\mathbf{x})$. Hadamard matrices exist at every order $2^{k}$ (Sylvester), at $q+1$ for a prime power $q \equiv 3 \pmod 4$ (Paley type I), and at $2(q+1)$ for $q \equiv 1 \pmod 4$ (Paley type II), and the Kronecker product $H_a \otimes H_b$ is Hadamard of order $ab$. Each coordinate of $H\mathbf{x}$ is a signed sum $\tfrac{1}{\sqrt n}\sum_j \pm x_j$, so the rotation spreads a coordinate outlier over all $n$ outputs and shrinks $r_t$ in Eq.~(\ref{eq:act-quant})~\citep{ashkboos2024quarot,tseng2024quipsharp,li2025svdquant}. A component along $\mathbf{1}$ is the exception, since $H_n\mathbf{1}$ is the vector of row sums of $H_n$ and spreads that component uniformly only when all row sums are equal. Hadamard matrices with equal row sums are called regular~\citep{seberry1992hadamard}.

\begin{lemma}[Regular Hadamard matrices]
\label{lem:regular}
If all row sums of a Hadamard matrix $H_n$ with $n > 1$ equal $c$, then $n = 4u^{2}$ and $c = \pm 2u$ for an integer $u$. The Sylvester matrix $H_{2^k}$ is not regular for $k \ge 1$, and $H_a \otimes H_b$ is regular only if both factors are (proof in App.~\ref{sec:analysis}).
\end{lemma}

\section{Method}
\label{sec:method}


\subsection{Motivation: DC energy in DiT activations}
\label{sec:motivation}

\begin{figure}[!ht]
\centering
\includegraphics[width=0.8\textwidth]{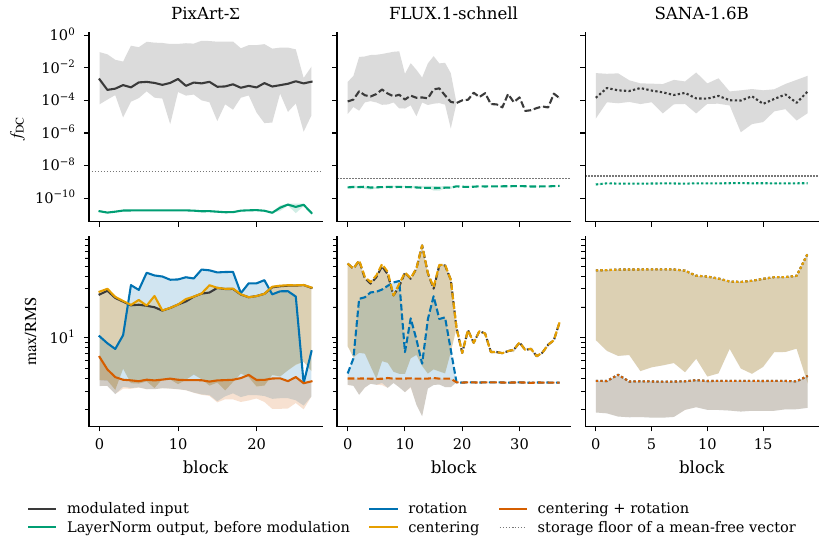}
\caption{\textbf{DC energy and per-token range across blocks.} Top: DC energy fraction $f_{\mathrm{DC}}$ of the modulated input and of the LayerNorm output. Bottom: per-token $\max/\mathrm{RMS}$ under identity, rotation only, centering only, and centering + rotation. Median over tokens at the highest-$\sigma$ step, band = min/max over layer types.}
\label{fig:dc-energy}
\end{figure}

\paragraph{The DC component and its origin.}
For a token $\mathbf{x}_t \in \mathbb{R}^{n}$ with channel mean $\bar{x}_t = \tfrac{1}{n}\mathbf{1}^{\top}\mathbf{x}_t$, we call $\bar{x}_t\mathbf{1}$ its DC component and measure its share of the token energy by
\begin{equation}
f_{\mathrm{DC}}(t) = \frac{n\,\bar{x}_t^{2}}{\|\mathbf{x}_t\|_2^{2}} \in [0,1].
\label{eq:fdc}
\end{equation}
In a DiT block, the QKV and first MLP projections receive the adaLN output $\mathbf{x}_t = (1+\boldsymbol{\gamma})\odot\tilde{\mathbf{x}}_t + \boldsymbol{\beta}$, where $\tilde{\mathbf{x}}_t$ is the LayerNorm output and the conditioning vectors $(\boldsymbol{\gamma},\boldsymbol{\beta})$, which depend on the timestep and prompt, are shared by all tokens. Because $\tilde{\mathbf{x}}_t$ has zero channel mean, the modulation reintroduces one, $\bar{x}_t = \bar{\beta} + \tfrac{1}{n}\boldsymbol{\gamma}^{\top}\tilde{\mathbf{x}}_t$ with $\bar{\beta} = \tfrac{1}{n}\mathbf{1}^{\top}\boldsymbol{\beta}$, whose second term varies per token and therefore cannot be folded into a bias. The second MLP projection receives post-activation features, which also have a nonzero mean. Accordingly, $f_{\mathrm{DC}}$ is nonzero on the modulated input, whereas on the LayerNorm output it stays at the floor set by storage precision (Fig.~\ref{fig:dc-energy}, top).

\paragraph{Rotation does not remove it.}
Because $HP$ preserves norms ($P$ zero-pads to width $n'$, Sec.~\ref{sec:activation}), the DC component keeps its energy $n\bar{x}_t^{2}$ after rotation and maps to $\bar{x}_t\,HP\mathbf{1}$, whose coordinates are the row sums of $HP$. For a factorized rotation $H = (H_m \otimes H_{2^k})/\sqrt{n'}$ (Sec.~\ref{sec:activation}), the Sylvester factor sends $\mathbf{1}$ onto its first coordinate, so at least a fraction $n/n'$ of this energy falls on $m$ coordinates, and all of it when $2^k$ divides $n$. A power-of-two rotation ($m = 1$) therefore places it in a single coordinate of magnitude $|\bar{x}_t|\,n/\sqrt{n'}$. Uniform spreading requires a regular Hadamard matrix (Lemma~\ref{lem:regular}). Even then it shifts every coordinate by $\pm\bar{x}_t$, so the range still grows with $|\bar{x}_t|$.

\paragraph{Consequence for 4-bit activations.}
The per-token step $\Delta_t \propto \max_j |(HP\mathbf{x}_t)_j|$ is then set by the DC coordinate, so the incoherent bulk of the token is quantized on a grid sized for the DC spike. Consistent with this, rotation alone leaves the per-token $\max/\mathrm{RMS}$ ratio at the level of the unrotated token, whereas centering before rotation lowers it to the level that incoherence predicts for the centered part (Fig.~\ref{fig:dc-energy}, bottom).

\subsection{Activation path}
\label{sec:activation}

\paragraph{Centering and exact re-injection.}
Each token is centered over its $n$ native channels, $\mathbf{x}_t^{c} = \mathbf{x}_t - \bar{x}_t\mathbf{1}$, at a cost of $O(n)$ per token and one scalar stored alongside the per-token scale, and the fused kernel of App.~\ref{app:algorithm} computes the mean in the same launch as the range. Because the layer is linear, the removed component re-enters the output exactly through a rank-1 term,
\begin{equation}
W\mathbf{x}_t = W\mathbf{x}_t^{c} + \bar{x}_t\,\mathbf{w}_{\mathbf{1}}, \qquad \mathbf{w}_{\mathbf{1}} = W\mathbf{1} \in \mathbb{R}^{d_{\mathrm{out}}}.
\label{eq:reinject}
\end{equation}
The row-sum vector $\mathbf{w}_{\mathbf{1}}$ is computed once in FP32 from the unquantized weight and stored at the model's compute precision, so the DC component bypasses both quantizers.

\begin{figure}[t]
\centering
\begin{minipage}[c]{0.5\textwidth}
  \centering
  \includegraphics[width=\linewidth]{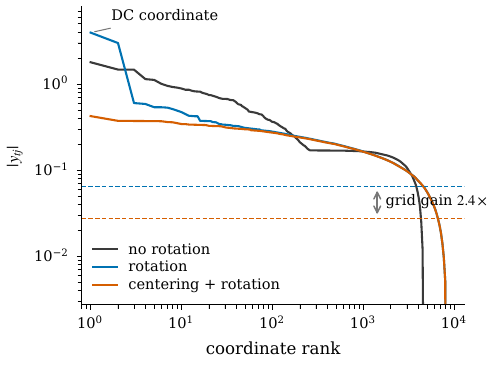}
\end{minipage}\hfill
\begin{minipage}[c]{0.46\textwidth}
  \caption{\textbf{Effect of centering on one token.} Sorted coordinate magnitudes after rotation, with and without centering; dashed levels mark the resulting INT4 step. Centering removes the DC coordinate, which otherwise sets the per-token range.}
  \label{fig:sorted-mag}
\end{minipage}
\vspace{-8pt}
\end{figure}

\paragraph{Rotation.}
The centered token is rotated by a fixed orthogonal matrix $H = (H_m \otimes H_{2^k})/\sqrt{n'}$ of order $n' = m\,2^k \ge n$, where $H_{2^k}$ is a Sylvester matrix and $H_m$ is a Paley matrix of order $m \in \{12, 36, 44\}$ or $m = 1$. When $n$ factors in this form, the rotation runs at the native width, and otherwise $P$ zero-pads the token to the smallest admissible $n'$. The factorization applies the rotation as two small transforms in $O(n'(m + 2^k))$ operations per token. Because centering precedes padding, $\mathbf{1}_{n'}^{\top}P\mathbf{x}_t^{c} = 0$ and the padded lanes are exactly zero, so the DC component is absent from the rotated token for every admissible $n'$, whatever the row sums of $H$ (Fig.~\ref{fig:sorted-mag}).

\paragraph{Scale and grid.}
The rotated token $HP\mathbf{x}_t^{c}$ is quantized to a symmetric $b$-bit grid with one scale per token. We estimate its spread as $\hat{\sigma}_t$, the empirical $\rho$-quantile of the magnitudes $|(HP\mathbf{x}_t^{c})_j|$ divided by the half-normal quantile $\Phi^{-1}((1+\rho)/2)$, and set $\alpha_t = \gamma^{\star}\hat{\sigma}_t / (2^{b-1}-1)$, where $\gamma^{\star} = 2.56$ minimizes the combined clipping and rounding error of a 4-bit grid for Gaussian input \citep{banner2019aciq}. Both $\rho = 0.9995$ and $\gamma^{\star}$ are fixed across models and layers, and $\hat{\sigma}_t$ and $\bar{x}_t$ are computed on the current token at inference.

\paragraph{Forward pass.}
One quantized linear layer computes
\begin{equation}
\label{eq:forward}
\mathbf{y}_t = \widehat{W}_{R}\,Q_a\!\big(HP\mathbf{x}_t^{c}\big) + \bar{x}_t\,\mathbf{w}_{\mathbf{1}} + \mathbf{b},
\qquad \widehat{W}_{R} \approx W P^{\top} H^{\top} \in \mathbb{R}^{d_{\mathrm{out}}\times n'},
\end{equation}
where the first term multiplies the 4-bit rotated weight of Sec.~\ref{sec:weights} with the 4-bit rotated token, and the rank-1 term and bias are fused into the epilogue of this matrix multiplication at compute precision (App.~\ref{app:algorithm}).

\begin{figure}[!ht]
\centering
\IfFileExists{figures/layer_dataflow.pdf}
  {\includegraphics[width=\textwidth]{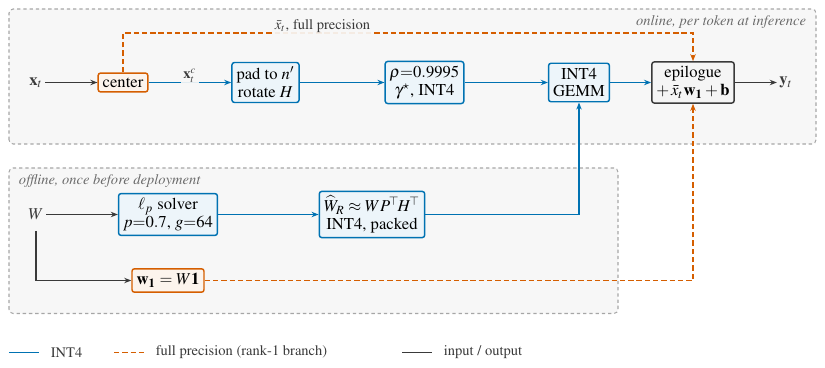}}
  {\fbox{\parbox[c][0.30\textheight][c]{\textwidth}{\centering \ph\ \texttt{figures/layer\_dataflow.pdf}\\[2pt]
   \footnotesize build it with \texttt{figures/src/layer\_dataflow.tex}}}}
\caption{\textbf{One quantized linear layer in CentriQ.} Online per-token path (top) and offline weight path (bottom). The per-token mean $\bar{x}_t$ and the row-sum vector $\mathbf{w}_{\mathbf{1}}$ form a full-precision rank-1 branch that bypasses both quantizers.}
\label{fig:dataflow}
\end{figure}

\paragraph{Offset invariance.}
Centering turns the DC component into an exact rank-1 term, so the quantized layer of Eq.~(\ref{eq:forward}) satisfies
\begin{equation}
\mathbf{y}(\mathbf{x}_t + c\mathbf{1}) = \mathbf{y}(\mathbf{x}_t) + c\,\mathbf{w}_{\mathbf{1}} \qquad \text{for every } c \in \mathbb{R},
\label{eq:invariance}
\end{equation}
and its deviation from $W\mathbf{x}_t + \mathbf{b}$ depends on $\mathbf{x}_t$ only through $\mathbf{x}_t^{c}$. The scale, the codes, and the quantization error therefore contain no terms that depend on the row sums of $H$, so the rotation can be chosen for mixing and width alone.

\subsection{Weight path}
\label{sec:weights}

The weight quantizer operates offline on the rotated weight $W_{R} = WP^{\top}H^{\top} \in \mathbb{R}^{d_{\mathrm{out}}\times n'}$, split into groups $W_{g}$ of $g = 64$ consecutive input columns. Row $i$ of group $g$ reconstructs as $s_i\,\sigma_{ig}\,(Q_{ig} - z_{ig})$, with 4-bit codes $Q_{g}$, a per-(row, group) scale $\boldsymbol{\sigma}_{g}$ fixed at its min/max initialization, a per-(row, group) shift $\mathbf{z}_{g}$, and a per-row scale $\mathbf{s}$ shared across the groups of a row. Since $s_i\sigma_{ig}$ is folded into a single 16-bit scale at export, the stored format matches Eq.~(\ref{eq:weight-quant}). The codes and the remaining parameters solve
\begin{equation}
\min_{\mathbf{s},\,\mathbf{z},\,Q}\ \sum_{g} \big\| W_{g} - \mathrm{diag}(\mathbf{s})\,\big[(Q_{g} - \mathbf{z}_{g})\odot\boldsymbol{\sigma}_{g}\big] \big\|_{p}^{p}, \qquad p = 0.7,\quad Q_{g} \in \{0,\dots,15\}^{d_{\mathrm{out}}\times g},
\label{eq:weight-objective}
\end{equation}
where $\|E\|_p^p = \sum_{ij}|E_{ij}|^p$. We call this weight quantizer RoDS and fix $p$ and $g$ across models and layers.

\paragraph{Why a robust $\ell_p$ objective.}
Min/max scaling, the $\ell_\infty$ criterion of RTN, and $\ell_2$ scale fits both let the largest entries of a group set the scale (Sec.~\ref{sec:prelim}). Under $\ell_p$ with $p < 1$, the marginal cost $p|r|^{p-1}$ of a residual decreases with $|r|$, so a few large entries are fitted poorly at small cost while the dense mode of the group sets the scale. Since DiT weight groups are heavy-tailed, a min/max scale stretches the grid over the tail and wastes resolution on the mode, whereas the $\ell_p$ fit resolves the mode. Tab.~\ref{tab:weight-only} compares the $\ell_p$ fit with min/max scaling (RTN), NF4, and HQQ.

\paragraph{Solver.}
We solve the nonconvex, nonsmooth problem in Eq.~(\ref{eq:weight-objective}) by nested alternation. The outer loop fits $\mathbf{s}$ with the codes and shifts frozen. Because the residual is linear in $\mathbf{s}$, each row reduces to an $\ell_p$ regression, which we solve by iteratively reweighted least squares (IRLS) with weights $(\epsilon + r_{ij}^{2})^{(p-2)/2}$ on the current residuals. The inner loop refits $(\mathbf{z}_{g}, Q_{g})$ on the preconditioned tile $W_{g}/\mathbf{s}$ by half-quadratic splitting, where an auxiliary residual $E_{g}$ decouples the $\ell_p$ term from the reconstruction,
\begin{equation}
\min_{\mathbf{z},\,Q,\,E}\ \sum_{g} \Big( \|E_{g}\|_{p}^{p} + \frac{\beta}{2}\,\big\| E_{g} - \big(W_{g}/\mathbf{s} - (Q_{g} - \mathbf{z}_{g})\odot\boldsymbol{\sigma}_{g}\big) \big\|_{F}^{2} \Big).
\label{eq:hqs}
\end{equation}
Each inner iteration rounds the scaled target $W_{g}/(\mathbf{s}\,\boldsymbol{\sigma}_{g}) + \mathbf{z}_{g}$ to the grid, shrinks $E_{g}$ with the generalized shrinkage of HQQ~\citep{badri2023hqq} as a one-step approximation of the proximal operator of $\|\cdot\|_p^p$, and sets $\mathbf{z}_{g}$ in closed form as the per-row mean offset between the codes and the residual-corrected target. Since $\mathbf{s}$ is initialized to $\mathbf{1}$ and $\boldsymbol{\sigma}_{g}$ stays at its min/max initialization, the first inner iteration coincides with RTN, and $\mathbf{s}$ is the only scale fitted under $\ell_p$. The penalty $\beta$ is fixed, and the iteration counts and tolerances are listed in App.~\ref{app:solver}.

\section{Results}
\label{sec:results}
We evaluate CentriQ on PixArt-$\Sigma$, SANA-1.6B, and FLUX.1-schnell at $1024^2$ resolution, using MJHQ-5k and sDCI following the SVDQuant protocol, plus GenEval on FLUX.1-schnell. All methods are run with identical prompts, seeds, and metric code, and design choices were made on a disjoint development split. Full details of models, samplers, metrics, baselines, and hardware are given in Appendix~\ref{app:setup}.


Tab.~\ref{tab:main-w4a4} compares W4A4 methods on PixArt-$\Sigma$, FLUX.1-schnell, and SANA-1.6B. Without calibration, CentriQ matches calibrated SVDQuant: relative to each method's 16-bit reference, FID and ImageReward are comparable on all three models, and CentriQ attains the lowest LPIPS. On SANA-1.6B, CentriQ stays within 0.2 FID of the BF16 model on both benchmarks. SVDQuant attains a higher PSNR on PixArt-$\Sigma$ and SANA-1.6B, whereas CentriQ attains a lower LPIPS, which weights perceptual structure over pixel-wise intensity. Both methods reach an FID below their 16-bit reference on FLUX.1-schnell, which lies within the noise of FID at 5k samples~\citep{chong2020effectively}.


RTN, NF4, and HQQ collapse on PixArt-$\Sigma$ and degrade substantially on FLUX.1-schnell and SANA-1.6B, so among the evaluated methods CentriQ is the only calibration-free one that produces usable images on all three models. CentriQ achieves this with the same grouped 4-bit weight format as RTN and HQQ.

\definecolor{oursrow}{RGB}{232,240,254}
\newcommand{\best}[1]{\textbf{#1}}
\newcommand{\second}[1]{\underline{#1}}

\begin{table}[!ht]
\centering
\tiny
\setlength{\tabcolsep}{4pt}
\caption{%
\textbf{Quantitative comparison of W4A4 quantization.} Best quantized result in \textbf{bold}, second best \underline{underlined}. SVDQuant\pub{} rows are taken from \citet{li2025svdquant}.
}
\label{tab:main-w4a4}
\resizebox{\textwidth}{!}{%
\begin{tabular}{l c cccc cccc}
\toprule
 & & \multicolumn{4}{c}{\textbf{MJHQ}} & \multicolumn{4}{c}{\textbf{sDCI}} \\
\cmidrule(lr){3-6}\cmidrule(lr){7-10}
Method & Calib.
       & FID $\downarrow$ & IR $\uparrow$ & PSNR $\uparrow$ & LPIPS $\downarrow$
       & FID $\downarrow$ & IR $\uparrow$ & PSNR $\uparrow$ & LPIPS $\downarrow$ \\
\midrule
\multicolumn{10}{l}{\textit{PixArt-}$\Sigma$} \\
\midrule
FP16             & \xmark & 16.7 & 0.952 & --  & --  & 24.9 & 0.973 & --  & --  \\
RTN              & \xmark & 481.5 & -2.273 & 6.14 & 0.995 & 470.1 & -2.277 & 6.60 & 0.957 \\
NF4              & \xmark & 479.6 & -2.273 & 6.16 & 0.997 & 468.6 & -2.277 & 6.46 & 0.966 \\
HQQ              & \xmark & 462.6 & -2.273 & 6.37 & 0.991 & 456.0 & -2.277 & 6.58 & 0.963 \\
SVDQuant\pub    & \cmark & \second{19.2} & \second{0.878} & \best{17.6} & \second{0.323} & \second{25.9} & \second{0.918} & \best{16.5} & \second{0.352} \\
\rowcolor{oursrow}
\textbf{CentriQ} & \xmark & \best{19.1} & \best{0.891} & \second{16.15} & \best{0.278} & \best{25.2} & \best{0.950} & \second{14.96} & \best{0.299} \\
\midrule
\multicolumn{10}{l}{\textit{FLUX.1-schnell}} \\
\midrule
BF16             & \xmark & 19.5 & 0.966 & --  & --  & 20.9 & 0.984 & --  & --  \\
RTN              & \xmark & 43.9 & -0.148 & 11.04 & 0.567 & 37.4 & 0.009 & 10.88 & 0.560 \\
NF4              & \xmark & 41.3 & -0.086 & 11.20 & 0.560 & 35.7 & 0.083 & 10.88 & 0.553 \\
HQQ              & \xmark & 40.9 & -0.084 & 11.29 & 0.558 & 34.2 & 0.120 & 10.93 & 0.551 \\
SVDQuant\pub    & \cmark & \best{18.3} & \best{0.951} & \second{18.3} & \second{0.258} & \best{20.1} & \best{0.979} & \second{17.2} & \second{0.260} \\
\rowcolor{oursrow}
\textbf{CentriQ} & \xmark & \second{18.8} & \second{0.939} & \best{18.40} & \best{0.173} & \second{20.6} & \second{0.970} & \best{17.29} & \best{0.173} \\
\midrule
\multicolumn{10}{l}{\textit{SANA-1.6B}} \\
\midrule
BF16             & \xmark & 16.0 & 1.097 & --  & --  & 22.4 & 1.061 & --  & --  \\
RTN              & \xmark & 33.3 & -0.330 & 12.02 & 0.497 & 43.4 & -0.703 & 9.29 & 0.621 \\
NF4              & \xmark & 33.3 & -0.457 & 11.93 & 0.503 & 43.3 & -0.788 & 10.67 & 0.551 \\
HQQ              & \xmark & 33.2 & -0.484 & 11.85 & 0.505 & 44.0 & -0.791 & 9.07 & 0.630 \\
SVDQuant\pub    & \cmark & \second{19.3} & \second{0.935} & \best{17.8} & \second{0.220} & \second{28.1} & \second{0.846} & \best{16.2} & \second{0.242} \\
\rowcolor{oursrow}
\textbf{CentriQ} & \xmark & \best{16.2} & \best{1.072} & \second{17.64} & \best{0.172} & \best{22.3} & \best{1.025} & \second{15.84} & \best{0.196} \\
\bottomrule
\end{tabular}%
}
\end{table}

\begin{figure}[!ht]
\centering
\includegraphics[width=\textwidth]{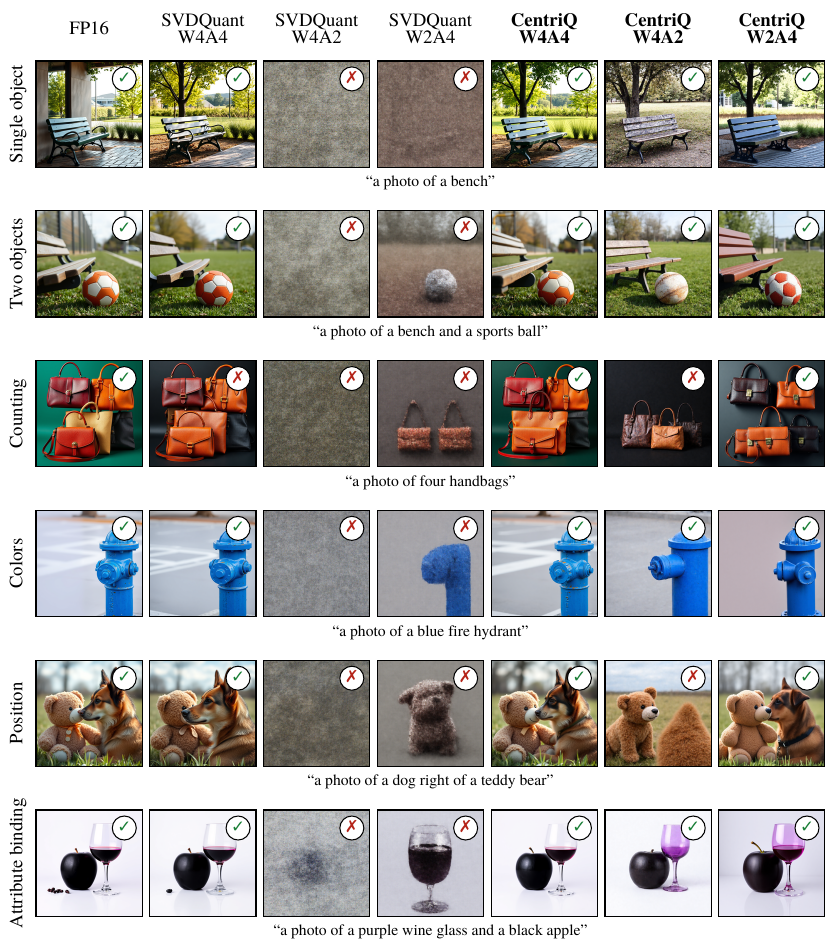}
\caption{\textbf{GenEval prompts on FLUX.1-schnell at W4A4, W4A2, and W2A4.} Images in a row share the prompt and seed, and the marks give the verdict of the GenEval detector.}
\label{fig:qual-geneval}
\end{figure}

Fig.~\ref{fig:qual-geneval} compares SVDQuant and CentriQ on GenEval prompts at W4A4, W4A2, and W2A4, and Figs.~\ref{fig:qual-w4a4-flux}, \ref{fig:qual-w4a4-pixart}, and~\ref{fig:qual-w4a4-sana} in the appendix compare them at W4A4 on all three models. All images in a row share the prompt, seed, scheduler, and step count, which is the same seed-level pairing that underlies the fidelity metrics in Tab.~\ref{tab:main-w4a4}.


\subsection{Latency results}
We measure the latency of CentriQ at W4A4 on FLUX.1-schnell on a single A100-80GB GPU at 1024$\times$1024 resolution and $4$ denoising steps. The deployed model uses a single-pass activation quantization kernel together with Nunchaku's 4-bit GEMMs. CentriQ generates an image in 1.69\,s, whereas the BF16 model takes 1.77\,s. The full calibration-free pipeline therefore runs slightly faster than the unquantized model as the time saved by the 4-bit matrix multiplication covers the cost of the added operations.

\section{Ablations}
\label{sec:ablations}

Tab.~\ref{tab:component-ablation} adds rotation, centering with exact re-injection, and percentile clipping one at a time to a baseline that uses the grouped $\ell_p$ weight quantizer and per-token activation quantization without any transform. The rotation-only row corresponds to a QuaRot-style data-free Hadamard rotation. Centering lowers FID from 25.65 to 18.46 at W4A4 and from 35.84 to 21.81 at W2A4. Percentile clipping is decisive at W4A2, where it lowers FID from 444.21 to 35.59, and changes W4A4 FID by 0.34, so a single clip setting serves all three bit-widths.

\begin{table}[!ht]
\centering
\footnotesize
\setlength{\tabcolsep}{5pt}
\caption{\textbf{Component ablation across bit-width on FLUX.1-schnell (MJHQ).}}
\label{tab:component-ablation}
\begin{tabular}{l cc cc cc}
\toprule
 & \multicolumn{2}{c}{W4A4} & \multicolumn{2}{c}{W4A2} & \multicolumn{2}{c}{W2A4} \\
\cmidrule(lr){2-3}\cmidrule(lr){4-5}\cmidrule(lr){6-7}
Configuration & FID $\downarrow$ & IR $\uparrow$ & FID $\downarrow$ & IR $\uparrow$ & FID $\downarrow$ & IR $\uparrow$ \\
\midrule
No transform                        & 40.62 & -0.067 & 437.41 & -2.270 & 363.14 & -2.244 \\
$+$ rotation (QuaRot-style)         & 25.65 & 0.768 & 455.30 & -2.271 & 35.84 & 0.371 \\
$+$ centering (exact rank-1 re-injection) & 18.46 & 0.961 & 444.21 & -2.274 & 21.81 & 0.744 \\
$+$ percentile clip (\textbf{CentriQ}) & 18.80 & 0.939 & 35.59 & 0.394 & 21.56 & 0.728 \\
\midrule
BF16                                & 19.45 & 0.966 & 19.45 & 0.966 & 19.45 & 0.966 \\
\bottomrule
\end{tabular}
\end{table}

\section{Discussion and limitations}
\label{sec:discussion}

The per-token mean that CentriQ removes originates in the adaLN modulation, which all three conditioning designs share, so one activation path transfers across PixArt-$\Sigma$, FLUX.1-schnell, and SANA-1.6B without per-model tuning. By Lemma~\ref{lem:regular}, none of the evaluated DiT widths admits a rotation that spreads this component uniformly, which makes explicit mean removal the direct remedy for Hadamard-based activation quantization in these models. Because the centered, rotated token contains no term that depends on the row sums of the rotation (Eq.~(\ref{eq:range-split}), App.~\ref{sec:analysis}), a single clip percentile, fixed once for all models and layers, reaches the quality of calibrated SVDQuant in Tab.~\ref{tab:main-w4a4}.

\paragraph{Limitations.} The 2-bit configurations are evaluated on FLUX.1-schnell with simulated quantization, because no fused kernel yet supports 2-bit activations. Latency is measured on A100 integer tensor cores, and transferring CentriQ to the microscaling FP4 formats (NVFP4, MXFP4) of newer GPUs remains to be evaluated. The analysis targets DiTs conditioned through adaLN, which covers the three evaluated families, and the gain from centering is expected to be smaller in architectures whose activations carry little per-token mean.

\section{Conclusion}
\label{sec:conclusion}

We showed that adaptive layer normalization places a per-token mean in DiT activations, which a Hadamard rotation concentrates onto a few coordinates that dominate the quantization range. CentriQ centers each token before rotation and restores the mean exactly through a rank-1 term, which makes the quantization range independent of the rotation, and it quantizes weights under a robust $\ell_p$ objective, so no step requires calibration data. At W4A4, CentriQ matches calibrated SVDQuant on all three models, and its centering and rotation add no net latency over BF16 inference. On FLUX.1-schnell, it also retains usable image quality at W2A4 and extends calibration-free quantization to 2-bit activations. A fused kernel for 2-bit activations and an extension to video diffusion transformers are natural next steps.

\bibliography{iclr2027_conference}
\bibliographystyle{iclr2027_conference}

\appendix
\section{Experimental setup}
\label{app:setup}
 
\paragraph{Models.} We evaluate CentriQ on PixArt-$\Sigma$, SANA-1.6B and FLUX.1-schnell. The three models use different attention designs (full self-attention, linear attention, and joint text-image attention in an MM-DiT), and each injects conditioning through adaLN shift terms, which produce the per-token mean that centering removes. All three generate $1024^2$ images: PixArt-$\Sigma$ and SANA-1.6B sample with 20 steps of DPM-Solver and Flow-DPM-Solver, respectively, at a guidance scale of 4.5, whereas FLUX.1-schnell samples with 4 flow-matching Euler steps at a guidance scale of 0.

\paragraph{Benchmarks.} Following the SVDQuant protocol, we generate one image per prompt for the 5,000 prompts of MJHQ-5k and the 5,000 prompts of sDCI. Each prompt has a fixed seed that all methods share, so the paired metrics compare outputs generated from the same initial noise. On FLUX.1-schnell, we additionally report GenEval, which scores compositional prompts using an object detector on 553 prompts with four images per prompt. All configuration choices, including the rotation layout and the clip percentile, were made on a disjoint 1{,}000-prompt development split of MJHQ. The evaluation split was scored once per final configuration, so no reported number informed a design decision.
 
\paragraph{Metrics.} FID and ImageReward measure image quality, with FID computed against the benchmark's reference images. PSNR and LPIPS measure fidelity to the 16-bit model by comparing each quantized output with the 16-bit output for the same prompt and seed. Both are computed on decoded 8-bit RGB images, and PSNR uses a peak value of 255.

\paragraph{Baselines.} We run RTN, NF4, and HQQ in our harness with identical prompts, seeds, and metric code, and in their rows of Tab.~\ref{tab:main-w4a4} activations are quantized per token without a transform. For Tab.~\ref{tab:main-w4a4}, we report SVDQuant from \citet{li2025svdquant}, whose MJHQ-5k and sDCI protocol we follow. For GenEval and the qualitative figures, we run SVDQuant from the Nunchaku wheel with smoothing factors calibrated on COCO captions, as in the original work.
 
\paragraph{Hardware and kernels.} All generation and latency measurements run on a single NVIDIA A100-80GB GPU. The 4-bit GEMMs use Nunchaku, which we extend with the centering and rotation operations. We measure end-to-end latency on FLUX.1-schnell, including the text encoders and the VAE decoder, with CUDA events as the mean over 6 images after 2 warm-up runs. Configurations below 4 bits for weights or activations use simulated quantization because no fused kernel supports these formats, so all latency figures refer to 4-bit weights and activations.

\section{Implementation details}
\label{app:implementation}

\subsection{Rotation construction}
\label{app:rotation}
CentriQ rotates each activation-quantized input with the factorized Hadamard matrix $H = (H_m \otimes H_{2^k})/\sqrt{n'}$ of order $n' = m\,2^k$, where $H_{2^k}$ is the Sylvester matrix and $H_m$ is a tabulated matrix with $m \in \{1, 12, 36, 44\}$ ($H_{36}$ Paley type II, $H_{44}$ Paley type I). When the native width $n$ admits such a factorization, the rotation runs at $n' = n$, and otherwise the centered token is zero-padded by $P = [I_n;\, 0]$ to the smallest admissible $n' > n$. The weight absorbs $H^{\top}$ offline as $W P^{\top} H^{\top}$, so the layer output is unchanged, and all matrices are deterministic. The Kronecker structure $(H_m \otimes H_{2^k})\,\mathrm{vec}(X) = \mathrm{vec}(H_{2^k} X H_m^{\top})$, with the token reshaped to $X \in \mathbb{R}^{2^k \times m}$, reduces the transform from $n'^2$ to $n'(m + 2^k)$ multiply-adds.

The factorization also explains why centering must precede the rotation. At the native width, the Sylvester factor maps $\mathbf{1}_{2^k}$ onto its first coordinate, so $H \mathbf{1}_n = \sqrt{2^k/m}\,(H_m \mathbf{1}_m) \otimes \mathbf{e}_1$ and the offset $\bar{x}_t \mathbf{1}_n$ lands with its full energy $n \bar{x}_t^2$ on at most $m$ of the $n'$ coordinates. Removing the per-token mean over the $n$ native coordinates eliminates this spike exactly, whereas centering after padding would leave a fraction $(n'-n)/n'$ of the offset energy in the operand.

\subsection{Fused kernel}
\label{app:algorithm}
The W4A4 GEMM extends the INT4 kernels of Nunchaku on an A100-SXM4-80GB and evaluates
\begin{equation}
\hat{\mathbf{y}}_t = \widetilde{W}\, Q_a\!\left(H P \mathbf{x}_t^{c}\right) + \bar{x}_t\, \mathbf{r} + \mathbf{b}, \qquad \mathbf{r} = W \mathbf{1}_n,
\label{eq:kernel}
\end{equation}
with $\widetilde{W} = Q_W(W P^{\top} H^{\top})$ and $\mathbf{r}$ precomputed from the full-precision weight in FP32. The prologue loads each token tile once, subtracts its mean, applies the factorized $H$ as tensor-core products, and quantizes to $\{-7, \dots, 7\}$ with one scale per token. Since weight codes are stored as $Q - 8$ with zero point $z' = z - 8$, the asymmetric grid leaves a correction $C = s z'$ per row and 64-channel group. Nunchaku's low-rank side branch absorbs this correction together with the rank-1 term and the bias through the activation columns $[\,\alpha_t \mathbf{p}_t \mid \bar{x}_t \mid 1\,]$ and weight columns $[\,-C \mid \mathbf{r} \mid \mathbf{b}\,]$, where $\mathbf{p}_t$ holds the group sums of the activation codes, so the integer main product is unchanged.
 
On FLUX.1-schnell ($1024^2$, 4 steps, batch 1), the native factorization lets the CUDA-graphed W4A4 transformer run without padding, and the full pipeline takes 1.69~s per image (Sec.~5.1). The W4A2 configurations use simulated quantization and carry no latency claim.

\subsection{Solver settings}
\label{app:solver}
The weight quantizer fits asymmetric 4-bit groups of 64 input channels in the rotated basis, with 16-bit scale and zero point in the model precision (4.5 bits per weight). Starting from the HQQ min-max initialization, the solver alternates IRLS updates of the per-row scale with half-quadratic splitting updates of the codes and zero point, while the group scale stays at its min-max value. Activations use a symmetric per-token grid whose scale follows a Gaussian model: $\hat{\sigma}_t = q_{0.9995}(\lvert HP\mathbf{x}_t^{c} \rvert)/\Phi^{-1}(0.99975)$ and $\alpha_t = \gamma^{\star}\hat{\sigma}_t / (2^{b-1}-1)$. Tab.~\ref{tab:solver} lists all settings, shared across models and layers.

\begin{table}[!ht]
\centering
\footnotesize
\setlength{\tabcolsep}{5pt}
\caption{\textbf{Quantizer and solver settings.}}
\label{tab:solver}
\begin{tabular}{l l}
\toprule
Setting & Value \\
\midrule
Weight grid, group & $\{0, \dots, 15\}$, $g = 64$ \\
Scale and zero point & FP16 (PixArt-$\Sigma$), BF16 (FLUX, SANA) \\
Objective exponent $p$ & 0.7 \\
Penalty $\beta$ & 10, fixed \\
Outer rounds / IRLS / proximal iterations & $\leq 3$ / $\leq 5$ / 20 \\
Stopping tolerance & $10^{-3}$, relative \\
Column factor & fixed to 1 (active only in the weight-only runs of Tab.~\ref{tab:weight-only}) \\
Activation grid & $\{-(2^{b-1}-1), \dots, 2^{b-1}-1\}$, per token, dynamic \\
Activation quantile / clip $\gamma^{\star}$ & 0.9995 / 2.5591 \\
\bottomrule
\end{tabular}
\end{table}

\section{Additional results}

\subsection{Sub-4-bit}
Tab.~\ref{tab:lowbit-flux} reports the corresponding fidelity metrics at sub-4-bit precision on FLUX.1-schnell. At W4A2, FID increases relative to W4A4, yet CentriQ keeps a positive ImageReward without any calibration data, which agrees with the compositional accuracy it retains in Tab.~\ref{tab:geneval-flux}.

\begin{table}[!ht]
\centering
\tiny
\setlength{\tabcolsep}{4pt}
\caption{%
\textbf{Sub-4-bit quantization of FLUX.1-schnell with CentriQ.} 
}
\label{tab:lowbit-flux}
\resizebox{\textwidth}{!}{%
\begin{tabular}{l cccc cccc}
\toprule
 & \multicolumn{4}{c}{\textbf{MJHQ}} & \multicolumn{4}{c}{\textbf{sDCI}} \\
\cmidrule(lr){2-5}\cmidrule(lr){6-9}
Precision
       & FID $\downarrow$ & IR $\uparrow$ & PSNR $\uparrow$ & LPIPS $\downarrow$
       & FID $\downarrow$ & IR $\uparrow$ & PSNR $\uparrow$ & LPIPS $\downarrow$ \\
\cmidrule(lr){2-5}\cmidrule(lr){6-9}
BF16 & 19.45 & 0.966 & --    & --    & 20.92 & 0.984 & --    & --    \\
\cmidrule(lr){2-5}\cmidrule(lr){6-9}
W4A4 & 18.80  & 0.939 & 18.40 & 0.173 & 20.6  & 0.970 & 17.29 & 0.173 \\
W4A2 & 35.59 & 0.394 & 13.62 & 0.457 & 28.61 & 0.683 & 12.85 & 0.429 \\
W2A4 & 21.56 & 0.728 & 14.35 & 0.367 & 22.21 & 0.873 & 13.48 & 0.350 \\
\bottomrule
\end{tabular}%
}
\end{table}

\subsection{GenEval}

Fidelity metrics measure agreement with the 16-bit output, not whether the quantized model still follows the prompt. We therefore evaluate compositional generation on FLUX.1-schnell with GenEval (Tab.~\ref{tab:geneval-flux}). For completeness, the table also includes results reported by the concurrent preprint OrbitQuant~\citep{lee2026orbitquant}. At W4A4, CentriQ preserves prompt adherence: its Overall score of 0.670 matches BF16 (0.667) and is close to our calibrated SVDQuant reproduction (0.682), although OrbitQuant and AdaTSQ score higher. Below four bits, the comparison favors CentriQ. At W2A4, the rotation and smoothing baselines collapse to near-zero scores, whereas CentriQ reaches 0.647, ahead of OrbitQuant (0.604) on every sub-task except Colors. At W4A2, CentriQ retains 0.573 and is the only calibration-free method reported in this regime. Because W2A4 outperforms W4A2 on every sub-task, activation precision limits compositional quality more than weight precision does, which supports the emphasis of Sec.~\ref{sec:method} on the activation quantizer.

\begin{table}[!ht]
\centering
\footnotesize
\setlength{\tabcolsep}{4pt}
\caption{%
\textbf{GenEval on FLUX.1-schnell.}
Scores on six compositional sub-tasks and Overall.
Rows marked \pub{} are taken from the concurrent preprint of \citet{lee2026orbitquant}, which we include for completeness; its FP16 reference matches ours in Overall score.
\textbf{Bold} and \underline{underlined} entries mark the best and second-best quantized result within each bit-width.
}
\label{tab:geneval-flux}
\begin{tabular}{l c ccc ccc c}
\toprule
Method & Calib.
 & Single & Two & Counting
 & Colors & Position & Color attr.
 & Overall $\uparrow$ \\
\midrule
BF16              & \xmark & 0.997 & 0.879 & 0.572 & 0.742 & 0.273 & 0.540 & 0.667 \\
\midrule
\multicolumn{9}{l}{\textit{W4A4}} \\
Q-DiT\pub         & \cmark & 0.741 & 0.424 & 0.378 & 0.418 & 0.073 & 0.208 & 0.373 \\
SmoothQuant\pub   & \cmark & 0.619 & 0.293 & 0.272 & 0.317 & 0.043 & 0.143 & 0.281 \\
QuaRot\pub        & \xmark & 0.819 & 0.543 & 0.472 & 0.519 & 0.118 & 0.275 & 0.458 \\
ViDiT-Q\pub       & \cmark & 0.888 & 0.586 & 0.516 & 0.585 & 0.130 & 0.268 & 0.495 \\
SVDQuant\pub      & \cmark & \underline{0.994} & \textbf{0.910} & 0.450 & 0.708 & 0.260 & 0.420 & 0.624 \\
AdaTSQ\pub        & \cmark & \textbf{0.997} & 0.894 & 0.622 & \underline{0.793} & 0.278 & 0.498 & 0.680 \\
OrbitQuant\pub    & \xmark & 0.991 & 0.881 & \textbf{0.706} & \textbf{0.803} & \textbf{0.323} & 0.512 & \textbf{0.703} \\
SVDQuant (ours)   & \cmark & \underline{0.994} & \underline{0.899} & \underline{0.625} & 0.747 & \underline{0.285} & \textbf{0.540} & \underline{0.682} \\
\rowcolor{oursrow}
\textbf{CentriQ}  & \xmark & \underline{0.994} & 0.871 & 0.622 & 0.726 & \underline{0.285} & \underline{0.523} & 0.670 \\
\midrule
\multicolumn{9}{l}{\textit{W4A2}} \\
\rowcolor{oursrow}
\textbf{CentriQ}  & \xmark & 0.934 & 0.677 & 0.553 & 0.710 & 0.135 & 0.428 & 0.573 \\
\midrule
\multicolumn{9}{l}{\textit{W2A4}} \\
QuaRot\pub     & \xmark & 0.006 & 0.000 & 0.000 & 0.000 & 0.000 & 0.000 & 0.001 \\
SmoothQuant\pub & \cmark & 0.000 & 0.000 & 0.000 & 0.000 & 0.000 & 0.000 & 0.000 \\
ViDiT-Q\pub     & \cmark & 0.006 & 0.000 & 0.000 & 0.000 & 0.000 & 0.000 & 0.001 \\
OrbitQuant\pub  & \xmark & \underline{0.972} & \underline{0.697} & \underline{0.575} & \textbf{0.766} & \underline{0.198} & \underline{0.420} & \underline{0.604} \\
\rowcolor{oursrow}
\textbf{CentriQ}  & \xmark & \textbf{0.988} & \textbf{0.826} & \textbf{0.653} & \underline{0.718} & \textbf{0.208} & \textbf{0.493} & \textbf{0.647} \\
\bottomrule
\end{tabular}
\end{table}

\subsection{Activation path ablations}
Tab.~\ref{tab:act-path} fixes the CentriQ activation path and varies the weight quantizer. With the activation path in place, all four quantizers produce usable images on FLUX.1-schnell and SANA-1.6B, where they collapse or degrade substantially without it (Tab.~\ref{tab:main-w4a4}). On PixArt-$\Sigma$, RoDS further lowers FID from 21.04 (HQQ) to 19.13.

\begin{table}[!ht]
\centering
\footnotesize
\setlength{\tabcolsep}{5pt}
\caption{\textbf{Activation path against weight quantizer at W4A4 (MJHQ).}}
\label{tab:act-path}
\begin{tabular}{l cc cc cc}
\toprule
 & \multicolumn{2}{c}{PixArt-$\Sigma$} & \multicolumn{2}{c}{FLUX.1-schnell} & \multicolumn{2}{c}{SANA-1.6B} \\
\cmidrule(lr){2-3}\cmidrule(lr){4-5}\cmidrule(lr){6-7}
Weight quantizer & FID $\downarrow$ & IR $\uparrow$ & FID $\downarrow$ & IR $\uparrow$ & FID $\downarrow$ & IR $\uparrow$ \\
\midrule
FP16  & 16.66 & 0.952 & 19.45 & 0.966 & 15.96 & 1.097 \\
\midrule
RTN   & 32.56 & 0.721 & 18.97   & 0.936   & 18.06   & 0.993   \\
NF4   & 24.40 & 0.834 & 19.19 & 0.926 & 17.97   & 0.942   \\
HQQ   & 21.04 & 0.866 & 18.92 & 0.940 & 16.30 & 1.073 \\
RoDS  & 19.13 & 0.891 & 18.80 & 0.939 & 16.19 & 1.072 \\
\bottomrule
\end{tabular}
\end{table}

\subsection{Weight path ablations}
Tab.~\ref{tab:weight-only} isolates the weight quantizer by keeping activations at 16 bits. At W3A16, the $\ell_p$ quantizer attains the lowest FID on all three models, and the margin is largest on PixArt-$\Sigma$, where it lowers FID from 34.36 (HQQ) to 29.04. ImageReward follows FID on PixArt-$\Sigma$ and FLUX.1-schnell. At W4A16, the four quantizers stay within 0.4 FID of each other on FLUX.1-schnell and SANA-1.6B, whereas on PixArt-$\Sigma$ the $\ell_p$ quantizer and HQQ lead RTN and NF4 by more than 2 FID.

\begin{table}[!ht]
\centering
\footnotesize
\setlength{\tabcolsep}{5pt}
\caption{\textbf{Weight-only quantization at W4A16 and W3A16 (MJHQ).}}
\label{tab:weight-only}
\begin{tabular}{l cc cc cc}
\toprule
 & \multicolumn{2}{c}{PixArt-$\Sigma$} & \multicolumn{2}{c}{FLUX.1-schnell} & \multicolumn{2}{c}{SANA-1.6B} \\
\cmidrule(lr){2-3}\cmidrule(lr){4-5}\cmidrule(lr){6-7}
Method & FID $\downarrow$ & IR $\uparrow$ & FID $\downarrow$ & IR $\uparrow$ & FID $\downarrow$ & IR $\uparrow$ \\
\midrule
FP16 & 16.66 & 0.952 & 19.45 & 0.966 & 15.96 & 1.097 \\
\midrule
\multicolumn{7}{l}{\textit{W4A16}} \\
RTN            & 23.04 & 0.823 & 19.02 & 0.968 & 16.32 & 1.107 \\
NF4            & 18.74 & 0.889 & 19.19 & 0.953 & 16.07 & 1.103 \\
HQQ            & 16.40 & 0.925 & 18.87 & 0.958 & 16.15 & 1.096 \\
\textbf{RoDS}  & 16.27 & 0.903 & 18.82 & 0.960 & 16.22 & 1.078 \\
\midrule
\multicolumn{7}{l}{\textit{W3A16}} \\
RTN            & 145.07 & -1.636 & 20.09 & 0.877 & 16.71 & 1.073 \\
HQQ            & 34.36  & 0.470  & 18.46 & 0.942 & 17.01 & 1.037 \\
\textbf{RoDS}  & 29.04  & 0.584  & 18.44 & 0.951 & 16.16 & 1.021 \\
\bottomrule
\end{tabular}
\end{table}

\begin{figure}[!ht]
\centering
\includegraphics[width=0.7\textwidth]{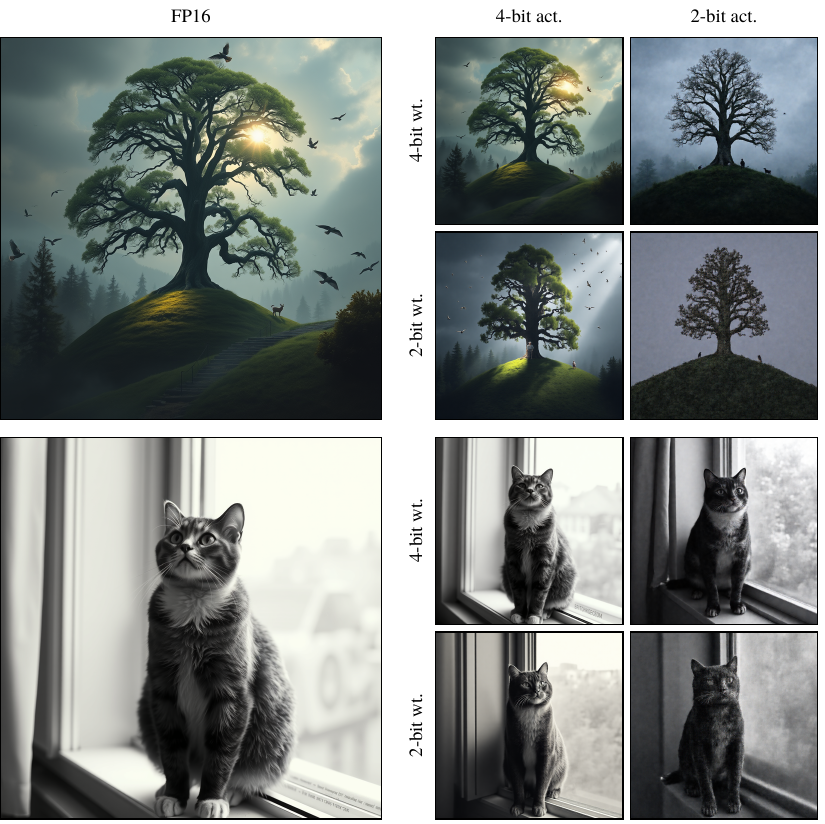}
\caption{\textbf{CentriQ on FLUX.1-schnell with 4-bit and 2-bit weights and activations.} The large image is the 16-bit output. In each grid, rows set the weight precision and columns set the activation precision, and every image shares the prompt and seed of the 16-bit output.}
\label{fig:qual-bits-a}
\end{figure}

\begin{figure}[!ht]
\centering
\includegraphics[width=0.7\textwidth]{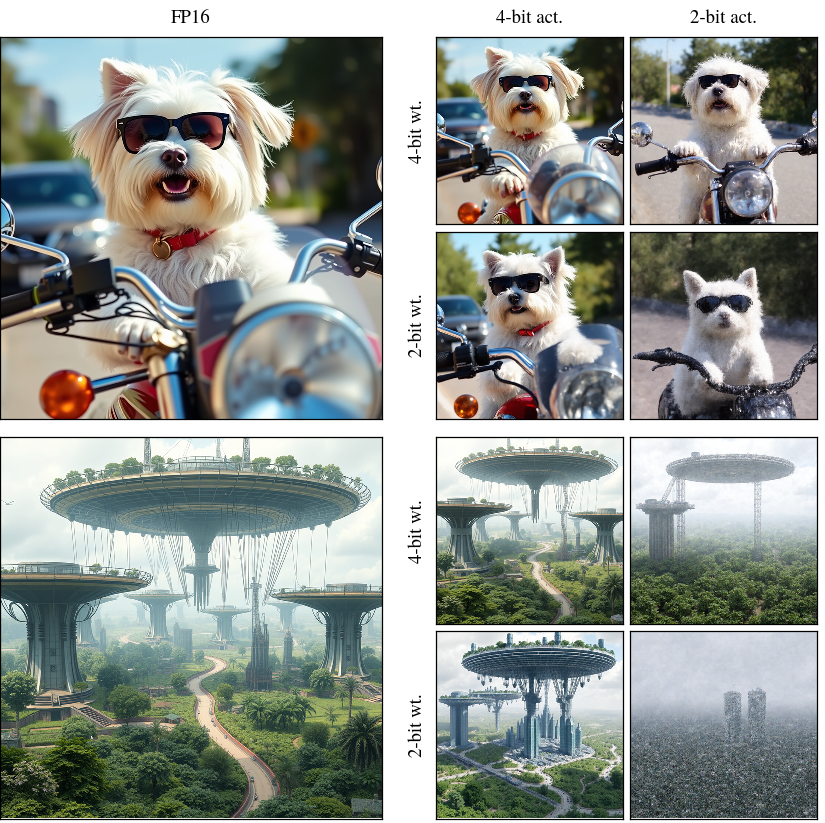}
\caption{\textbf{Further CentriQ samples on FLUX.1-schnell with 4-bit and 2-bit weights and activations.} The layout follows Fig.~\ref{fig:qual-bits-a}.}
\label{fig:qual-bits-b}
\end{figure}

\begin{figure}[!ht]
\centering
\includegraphics[width=0.7\textwidth]{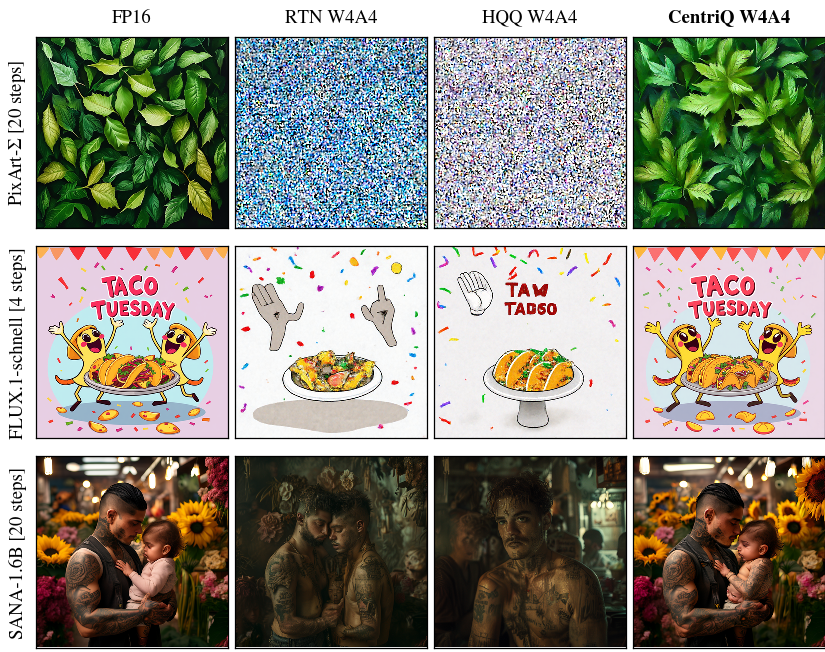}
\caption{\textbf{W4A4 with calibration-free quantizers on PixArt-$\Sigma$, FLUX.1-schnell, and SANA-1.6B.} RTN and HQQ produce noise on PixArt-$\Sigma$ and lose the content of the prompt on FLUX.1-schnell and SANA-1.6B, whereas CentriQ stays close to the 16-bit output.}
\label{fig:qual-w4a4-collapse}
\end{figure}

\begin{figure}[!ht]
\centering
\includegraphics[width=0.7\textwidth]{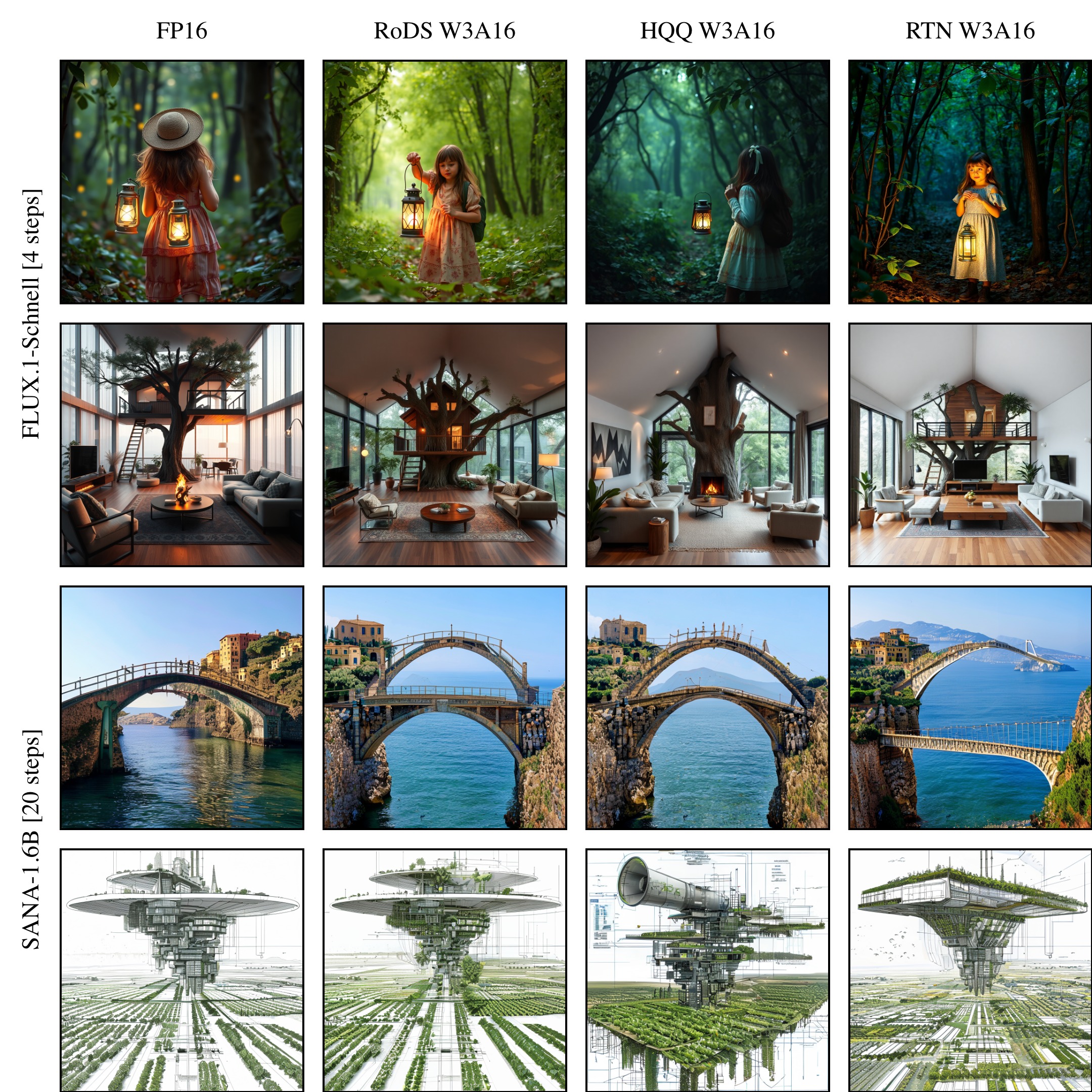}
\caption{\textbf{Weight-only W3A16 on FLUX.1-schnell and SANA-1.6B.} RoDS, our $\ell_p$ weight quantizer, compared with HQQ and RTN at 3-bit weights and 16-bit activations.}
\label{fig:qual-w3a16-flux-sana}
\end{figure}

\begin{figure}[!ht]
\centering
\includegraphics[width=0.7\textwidth]{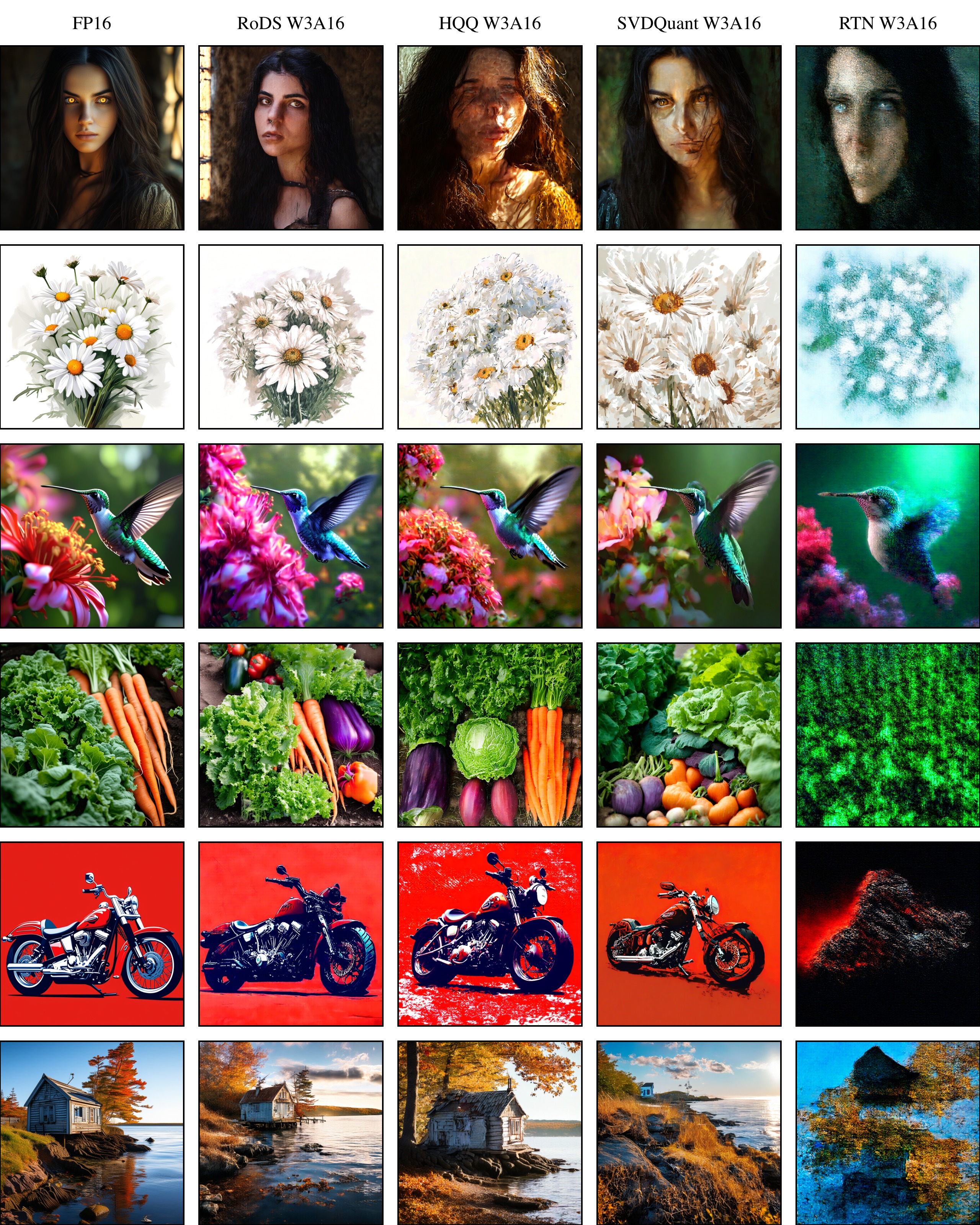}
\caption{\textbf{Weight-only W3A16 on PixArt-$\Sigma$.} RoDS, our $\ell_p$ weight quantizer, compared with HQQ, SVDQuant, and RTN at 3-bit weights and 16-bit activations.}
\label{fig:qual-w3a16-pixart}
\end{figure}

\begin{figure}[!ht]
\centering
\includegraphics[width=0.7\textwidth]{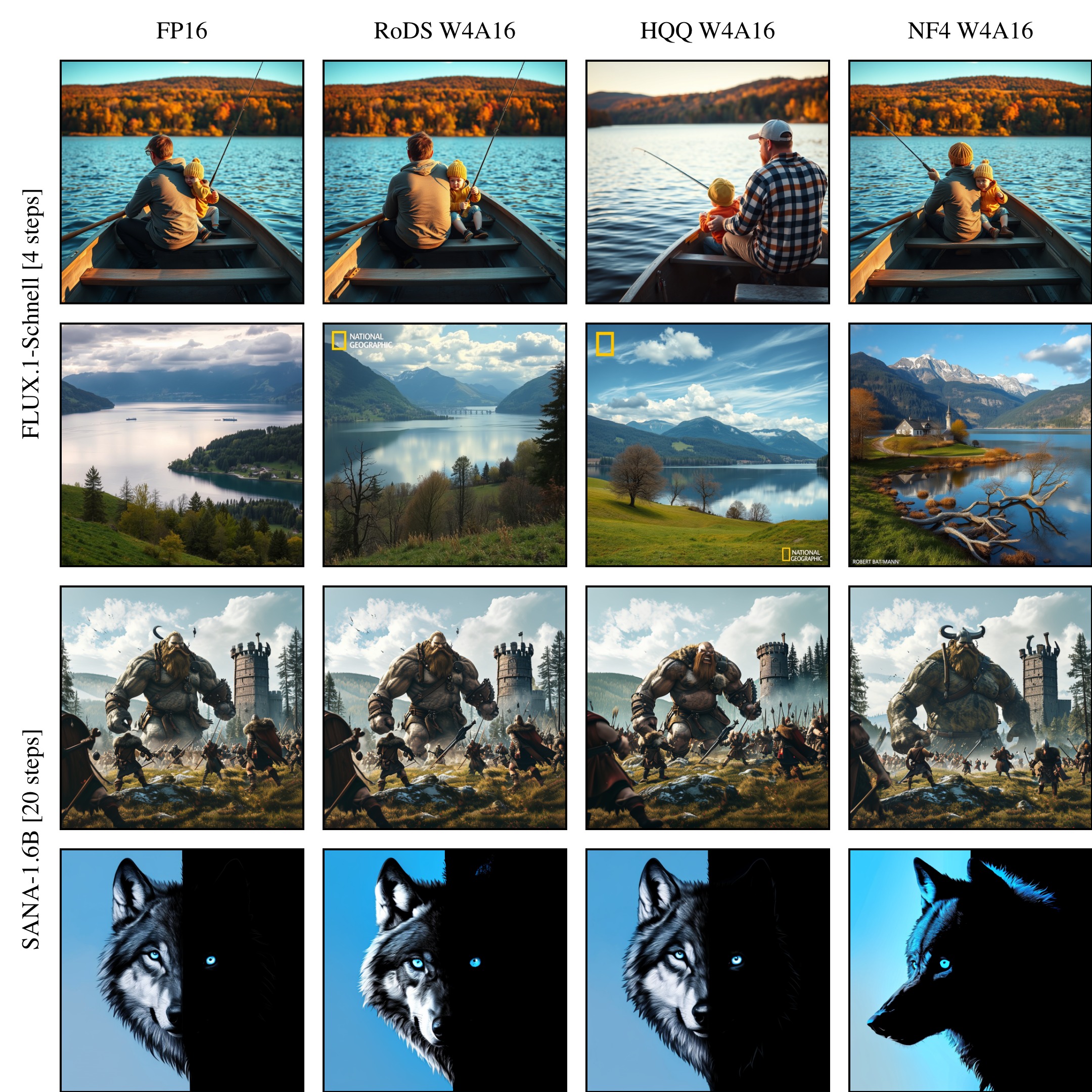}
\caption{\textbf{Weight-only W4A16 on FLUX.1-schnell and SANA-1.6B.} RoDS, our $\ell_p$ weight quantizer, compared with HQQ and NF4 at 4-bit weights and 16-bit activations.}
\label{fig:qual-w4a16-flux-sana}
\end{figure}

\begin{figure}[!ht]
\centering
\includegraphics[width=0.7\textwidth]{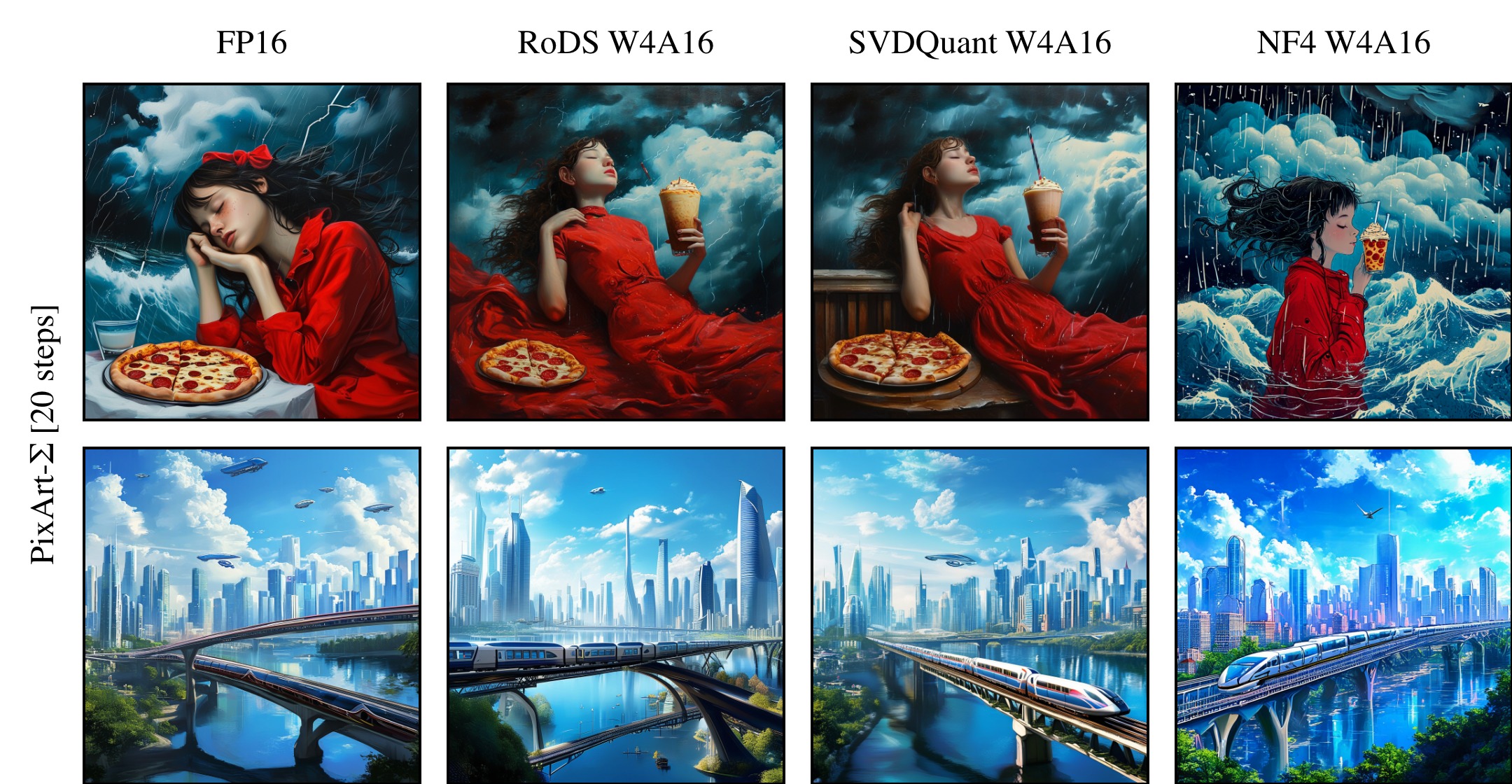}
\caption{\textbf{Weight-only W4A16 on PixArt-$\Sigma$.} RoDS, our $\ell_p$ weight quantizer, compared with SVDQuant and NF4 at 4-bit weights and 16-bit activations.}
\label{fig:qual-w4a16-pixart}
\end{figure}

\begin{figure}[!ht]
\centering
\includegraphics[width=0.7\textwidth]{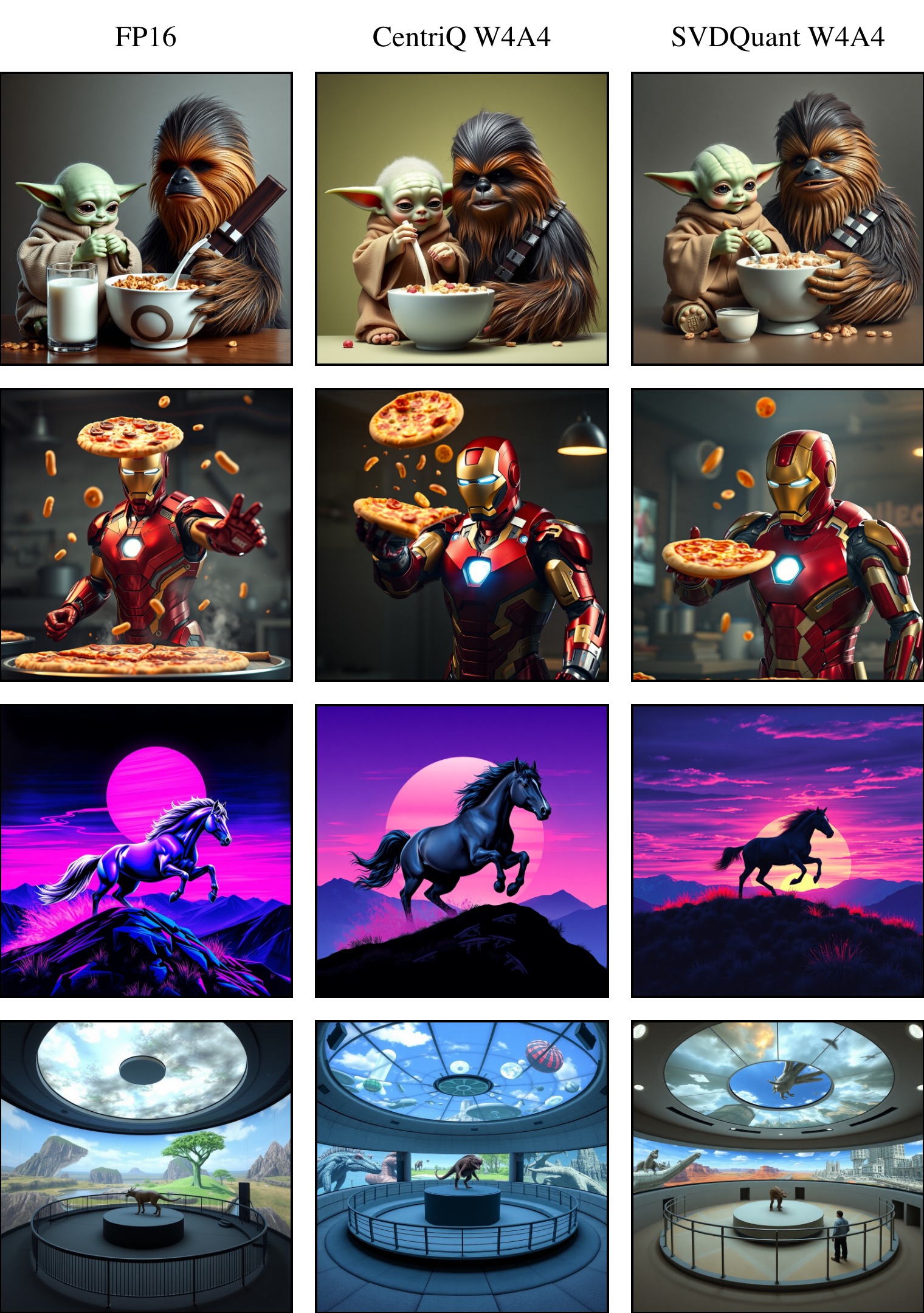}
\caption{\textbf{CentriQ and SVDQuant at W4A4 on FLUX.1-schnell.} Each row shares the prompt and seed of the BF16 output.}
\label{fig:qual-w4a4-flux}
\end{figure}

\begin{figure}[!ht]
\centering
\includegraphics[width=0.7\textwidth]{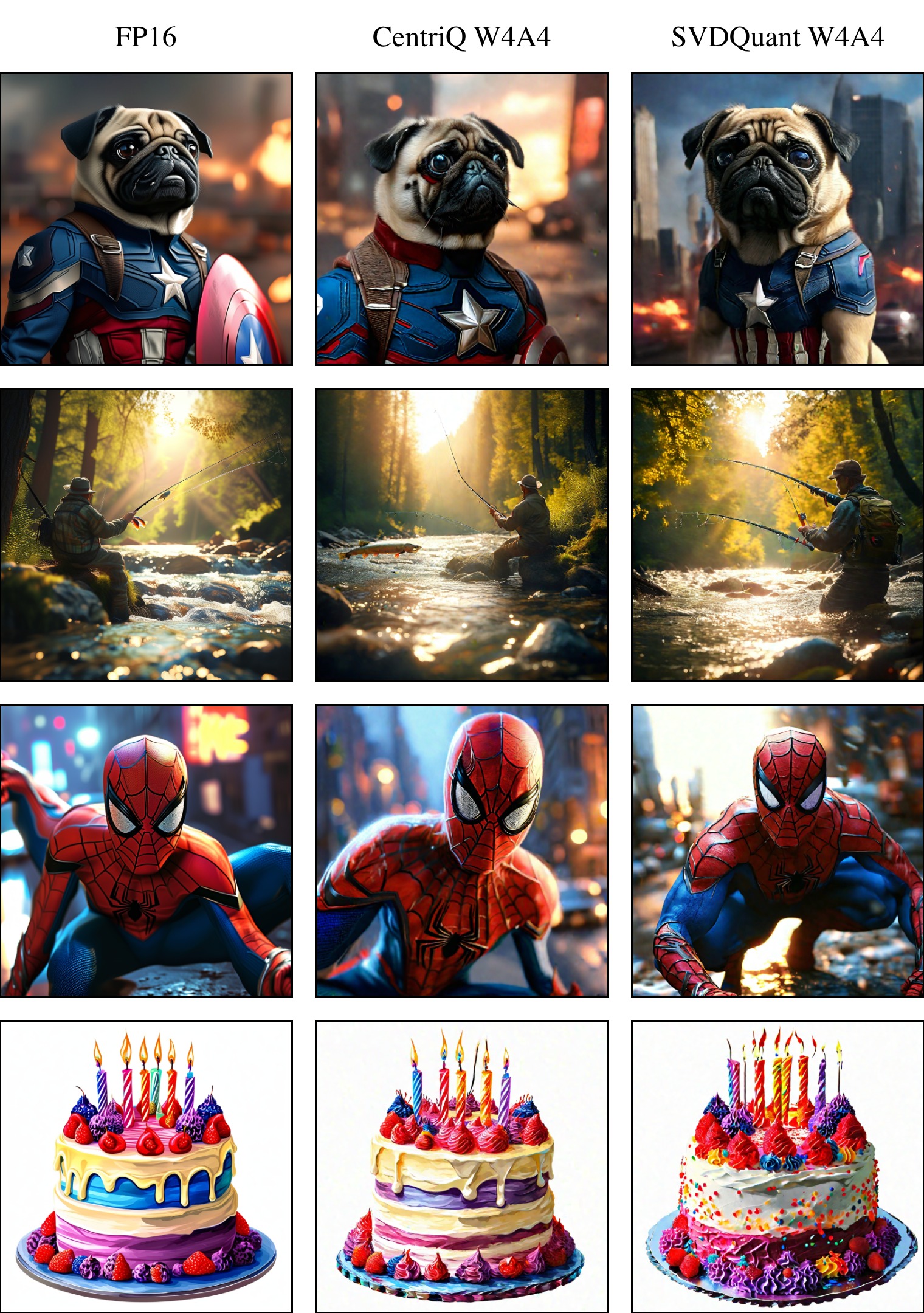}
\caption{\textbf{CentriQ and SVDQuant at W4A4 on PixArt-$\Sigma$.} Each row shares the prompt and seed of the FP16 output.}
\label{fig:qual-w4a4-pixart}
\end{figure}

\begin{figure}[!ht]
\centering
\includegraphics[width=0.7\textwidth]{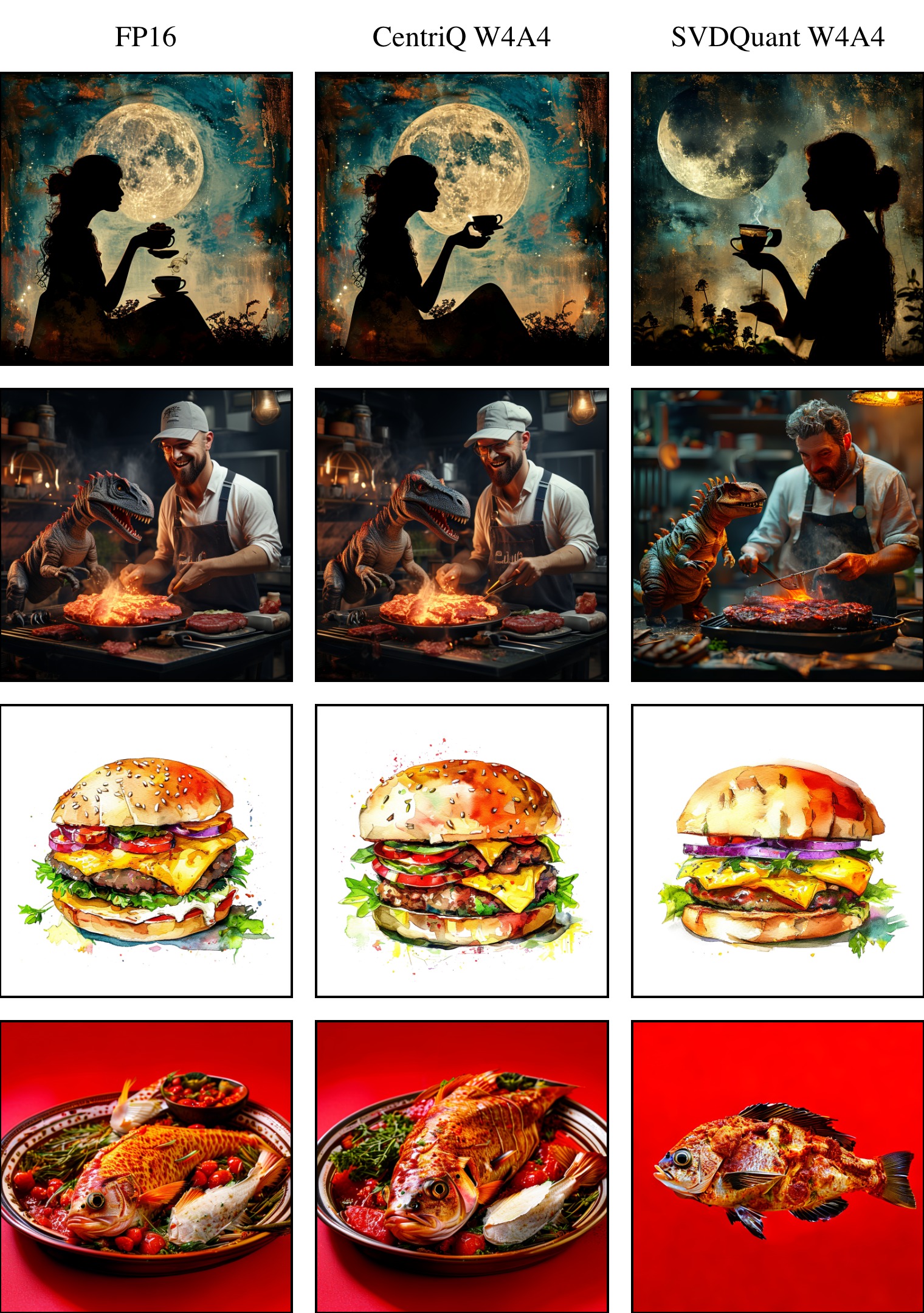}
\caption{\textbf{CentriQ and SVDQuant at W4A4 on SANA-1.6B.} Each row shares the prompt and seed of the BF16 output.}
\label{fig:qual-w4a4-sana}
\end{figure}

\subsection{Analysis}
\label{sec:analysis}

The forward pass in Eq.~(\ref{eq:forward}) contains two approximations, the weight quantizer $\widehat{W}_{R}$ and the activation quantizer $Q_a$, and one exact rewriting, the rank-1 branch. We first verify that the rewriting is exact, then show what centering removes from the range that $Q_a$ has to cover.

\paragraph{Exactness of the re-injection.}
The decomposition in Eq.~(\ref{eq:reinject}) is an algebraic identity in $W$ and $\mathbf{x}_t$, because $\mathbf{x}_t^{c} + \bar{x}_t\mathbf{1} = \mathbf{x}_t$ and $\mathbf{w}_{\mathbf{1}} = W\mathbf{1}$ by definition. Setting $\widehat{W}_{R} = WP^{\top}H^{\top}$ and $Q_a = \mathrm{id}$ in Eq.~(\ref{eq:forward}) therefore recovers $W\mathbf{x}_t + \mathbf{b}$ exactly, since $P^{\top}H^{\top}HP = I_n$. In finite-precision arithmetic the branch adds one rank-1 update in the epilogue, with $\mathbf{w}_{\mathbf{1}}$ stored at the model's compute precision. Consequently every deviation of CentriQ from the full-precision model is attributable to $\widehat{W}_{R}$ and $Q_a$, whereas the branch carrying the DC component contributes none.

\paragraph{What centering removes from the range.}
Because $HP$ is linear, the rotated token splits as
\begin{equation}
HP\mathbf{x}_t = HP\mathbf{x}_t^{c} + \bar{x}_t\,HP\mathbf{1},
\label{eq:range-split}
\end{equation}
where $HP\mathbf{1}$ is the vector of row sums of $HP$. The second term concentrates at least a fraction $n/n'$ of the DC energy on $m$ coordinates, one of which has magnitude at least $|\bar{x}_t|\,n/\sqrt{m n'}$, so without centering it lower-bounds the per-token range whenever the centered part does not cancel it, and Fig.~\ref{fig:sorted-mag} shows that it dominates the range in practice. Since the per-token step $\Delta_t$ is proportional to the range, this single coordinate coarsens the grid for all $n'$ coordinates of the token. Centering makes the second term identically zero, and since the padding is zero-padding applied after centering, $\mathbf{1}_{n'}^{\top}P\mathbf{x}_t^{c} = 0$ holds at the rotation's width. The rotated centered token $HP\mathbf{x}_t^{c}$ therefore contains no term that depends on the row sums of $H$, and the same holds for every quantity computed from it: the absmax range, the percentile scale of Sec.~\ref{sec:activation}, the quantized codes, and the quantization error. This is an identity for every token and every timestep and requires no distributional assumption. By Lemma~\ref{lem:regular}, no Hadamard matrix at the native hidden widths of the evaluated models has constant row sums, so without centering the row-sum term is present for every one of them, whereas with centering the rotation can be selected for mixing quality and width alone.

How small the remaining range $\|HP\mathbf{x}_t^{c}\|_\infty$ is depends on the centered token and on the mixing of the rotation. Under sub-Gaussian coordinates of $\mathbf{x}_t^{c}$ it concentrates at $O\!\big(\sqrt{\log n'}\,\|\mathbf{x}_t^{c}\|_2/\sqrt{n'}\big)$, so the per-token $\max/\mathrm{RMS}$ ratio of the rotated centered token grows only as $O(\sqrt{\log n'})$, consistent with the centering + rotation curves in Fig.~\ref{fig:dc-energy} (bottom).

\paragraph{Proof of Lemma~\ref{lem:regular}.}
If $H_n\mathbf{1} = c\mathbf{1}$, then $c^{2}n = \|H_n\mathbf{1}\|_2^{2} = \mathbf{1}^{\top}H_n^{\top}H_n\mathbf{1} = n^{2}$, so $c^{2} = n$. No Hadamard matrix of order 2 has this property because 2 is not a square, and every Hadamard order $n > 2$ is divisible by 4, so a square order satisfies $n = 4u^{2}$ with $c = \pm 2u$. One row of $H_{2^k}$ sums to $2^k$ and every other row sums to zero, so $H_{2^k}$ is not regular for $k \ge 1$. The row sums of $H_a \otimes H_b$ are the products $s_i t_j$ of the row sums $s_i$ of $H_a$ and $t_j$ of $H_b$. If all products equal $c$, then $c \neq 0$ by the first part, and fixing $j$ gives $s_i = c/t_j$ for every $i$, so all $s_i$ are equal, and the same argument shows that all $t_j$ are equal.

\subsection*{AI use statement}

We used large language models (LLMs) to edit the manuscript for grammar, wording, and consistency, and to check cross-references, table entries, and derivations for internal consistency. The authors reviewed and revised all AI-assisted text.
We take full responsibility for the content of this work, including text, claims, and artifacts produced with the aid of generative AI.

\section*{Reproducibility statement}

The paper specifies every component needed to reimplement CentriQ. Sec.~\ref{sec:activation} and App.~\ref{app:rotation} define the centering, the factorized Hadamard rotation, the padding rule, and the closed-form per-token scale, and Sec.~\ref{sec:weights} and App.~\ref{app:solver} give the $\ell_p$ objective, the solver, and all hyperparameters, which are shared across models and layers (Tab.~\ref{tab:solver}). App.~\ref{app:algorithm} describes how the rank-1 term and the zero-point correction are fused into the Nunchaku 4-bit GEMM. All models, prompt sets (MJHQ-5k, sDCI, GenEval), and metrics are public, and App.~\ref{app:setup} lists the samplers, step counts, guidance scales, seeding protocol, and the Nunchaku release used for SVDQuant.
 
\section*{Ethics statement}
 
CentriQ lowers the cost of running text-to-image models, which widens access for legitimate uses and for misuse alike. The quantized models inherit the safety behavior and biases of their 16-bit sources, and this work does not evaluate either. We collected no new data and ran no studies with human participants. All prompt sets come from public benchmarks and are used under their respective licenses.

\end{document}